\documentclass{sig-alternate-2013}

\newfont{\mycrnotice}{ptmr8t at 7pt}
\newfont{\myconfname}{ptmri8t at 7pt}
\let\crnotice\mycrnotice%
\let\confname\myconfname%

\permission{Permission to make digital or hard copies of all or part of this work for personal or classroom use is granted without fee provided that copies are not made or distributed for profit or commercial advantage and that copies bear this notice and the full citation on the first page. Copyrights for components of this work owned by others than ACM must be honored. Abstracting with credit is permitted. To copy otherwise, or republish, to post on servers or to redistribute to lists, requires prior specific permission and/or a fee. Request permissions from Permissions@acm.org.}
\conferenceinfo{KDD '15,}{August 10-13, 2015, Sydney, NSW, Australia}
\copyrightetc{\copyright~2015 ACM. ISBN \the\acmcopyr}
\crdata{978-1-4503-3664-2/15/08...\$15.00\\
DOI: http://dx.doi.org/10.1145/2783258.2788627}

\usepackage{framed}
\usepackage{multirow}
\usepackage{graphics}
\usepackage{url}
\usepackage{wrapfig}
\usepackage{algorithmic} 
\usepackage{algorithm} 
\usepackage{float} 
\usepackage{amsmath}
\usepackage{amssymb}
\usepackage{amsfonts}

\usepackage[table]{xcolor}
\usepackage{epsfig}
\usepackage{booktabs}

\definecolor{lightgray}{gray}{0.85}

\numberwithin{equation}{section}

\usepackage{subfig, float, epsfig, epstopdf, balance, algorithm}
\usepackage{helvet}
\usepackage{courier}
\usepackage{multirow}
\usepackage{mathtools}
\usepackage{hyperref}

\def\P{\mathbb{P}}

\usepackage{amsmath}

\usepackage[english]{babel}

\begin{document}

\title{E-commerce in Your Inbox: \\ Product Recommendations at Scale}

\numberofauthors{2}
\author{
\alignauthor
Mihajlo Grbovic, Vladan Radosavljevic, Nemanja Djuric, Narayan Bhamidipati\\
\affaddr{Yahoo Labs}\\
\affaddr{701 First Avenue, Sunnyvale, USA}\\
\email{\{mihajlo, vladan, nemanja, narayanb\}@yahoo-inc.com}\\
\alignauthor
Jaikit Savla, Varun Bhagwan, \\ Doug Sharp\\
\affaddr{Yahoo, Inc.}\\
\affaddr{701 First Avenue, Sunnyvale, USA}\\
\email{\{jaikit, vbhagwan, dsharp\}@yahoo-inc.com}\\
}

\maketitle

\begin{abstract}
In recent years online advertising has become increasingly ubiquitous and effective. Advertisements shown to visitors fund sites and apps that publish digital content, manage social networks, and operate e-mail services. Given such large variety of internet resources, determining an appropriate type of advertising for a given platform has become critical to financial success. Native advertisements, namely ads that are similar in look and feel to content, have had great success in news and social feeds. However, to date there has not been a winning formula for ads in e-mail clients. In this paper we describe a system that leverages user purchase history determined from e-mail receipts to deliver highly personalized product ads to Yahoo Mail users. We propose to use a novel neural language-based algorithm specifically tailored for delivering effective product recommendations, which was evaluated against baselines that included showing popular products and products predicted based on co-occurrence. We conducted rigorous offline testing using a large-scale product purchase data set, covering purchases of more than $29$ million users from $172$ e-commerce websites. Ads in the form of product recommendations were successfully tested on online traffic, where we observed a steady $9\%$ lift in click-through rates over other ad formats in mail, as well as comparable lift in conversion rates. Following successful tests, the system was launched into production during the holiday season of 2014.
\end{abstract}


\newpage
\section{Introduction}

Hundreds of millions of people around the world visit their e-mail inboxes daily, mostly to communicate with their contacts, although a significant fraction of time is spent checking utility bills, reading newsletters, and tracking purchases. To monetize this overwhelming amount of traffic, e-mail clients typically show display ads in a form of images alongside native e-mail content. Convincing users to exit the ``e-mail mode", characterized by a relentless focus on the task of dealing with their mail, in order to enter a mode where they are willing to click on ads is a challenging task. 
Effective personalization and targeting \cite{essex2009matchmaker}, where the goal is to find the best matching ads to be displayed for each individual user, is essential to tackling this problem, as ads need to be highly relevant to overcome user's inclination to focus narrowly on the e-mail task.
In addition to financial gains for online businesses \cite{riecken2000personalized}, proactively tailoring advertisements to the tastes of each individual consumer also leads to an improved user experience, and can help with increasing user loyalty and retention \cite{alba1997interactive}. 

Inbound e-mails are still insufficiently explored and exploited for the purposes of ad targeting, while arguably representing a treasure trove of monetizable data. A recent study \cite{grbovic2014many} showed that only $10\%$ of inbound volume represents human-generated e-mails. Furthermore, out of the remaining $90\%$ of traffic more than $22\%$ represents e-mails related to online shopping. Given that a significant percentage of overall traffic has commercial intent, a popular form of targeted advertising is mail retargeting (MRT), where advertisers target users who previously received e-mails from certain commercial web domains. These e-mails present a strong signal useful for targeting campaigns, as they give a broad picture about each customer's interests and relationship with commercial domains. A recent paper \cite{grbovic2014generating} proposed to make use of this vast potential and presented a clustering method to generate MRT rules, showing that such rules are more accurate than the ones generated by human experts.

\begin{figure*}[t]
\centering
{\includegraphics[width=1\textwidth]{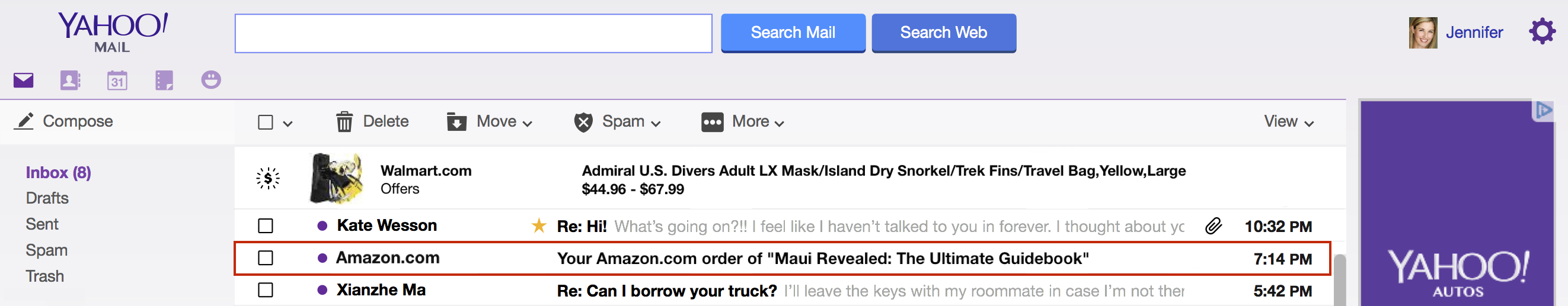}} 
\caption{Product recommendations in Yahoo Mail}
\label{fig:prod_ads}
\end{figure*}

However, in order to go beyond MRT rules that utilize only the fact that users and e-commerce sites communicated, advertisers require more detailed data such as purchased product name and price that are often part of e-mail body. E-mail clients have been working with e-commerce and travel domains to standardize how e-mails are formatted, resulting in schemas maintained by the {\it schema.org} community\footnote{http://schema.org/email, accessed June 2015}. With more and more e-commerce sites using standard schemas, e-mail clients can provide more personalized user notifications, such as package tracking\footnote{yahoomail.tumblr.com/post/107247385566/track-your-packages-now-in-the-yahoo-mail-app, accessed June 2015} and flight details\footnote{venturebeat.com/2014/10/22/yahoo-mail-now-tells-you-about-upcoming-flights-and-events, accessed June 2015}. 
In addition, e-mail receipt extraction brings monetization opportunity through product advertising to users based on their individual purchase history. Availability of purchase data from multiple commercial e-mail domains puts the e-mail provider in the unique position to be able to build better recommendation systems than those based on any one commercial e-mail domain alone. In particular, unlike e-commerce websites that make recommendations of type: ``Customers who bought $X$ also bought $Y$", e-mail providers can make recommendations of type: ``Customers who bought $X$ from vendor $V_1$ also bought $Y$ from vendor $V_2$", allowing much more powerful and effective targeting solutions.

In this paper we tell the story of an end-to-end development of product ads for Yahoo Mail. The effort included developing a product-level purchase prediction algorithm, capable of scaling to millions of users and products. To this end, we propose an approach that embeds products into real-valued, low-dimensional vector space using a neural language model applied to a time series of user purchases. As a result, products with similar contexts (i.e., their surrounding purchases) are mapped to vectors that are nearby in the embedding space. To be able to make meaningful and diverse suggestions about the next product to be purchased, we further cluster the product vectors and model transition probabilities between clusters. The closest products in the embedding space from the most probable clusters are used to form final recommendations for a product.

The product prediction model was trained using a large-scale purchase data set, comprising more than $280$ million purchases made by $29$ million users, involving $2.1$ million unique products. The model was evaluated on a held-out month, where we tested the effectiveness of recommendations in terms of yield rate. In addition, we evaluated several baseline approaches, including showing popular products to all users, showing popular products in various user groups (called {\it cohorts}, specified by user's gender, age, and location), as well as showing products that are historically frequently purchased after the product a user most recently bought. To mitigate the cold start problem, popular products in user's cohort were used as back-fill recommendations for users without earlier purchases. 

Empirical results show that the proposed product-level model is able to more accurately predict purchases than baseline approaches. In the experimental section we also share results of bucket tests on live traffic, where we compared the performance of the baselines in the same ad slot. Our method substantially improves key business metrics, measured in terms of clicks and conversions.
Moreover, our product-prediction technique was successfully implemented at large scale, tested in live buckets, and finally launched in production. The system is able to make near-real-time product recommendations with latency of less than $200ms$. In Figure \ref{fig:prod_ads} we show an example of our product recommender, where an offer from {\it walmart.com} is suggested after related purchase was made on {\it amazon.com} (marked with red). 

\section{Related work}
In this section we describe related work in mail-based advertising, as well as neural language models that motivated our product recommendation approach.

\subsection{Leveraging e-mail data in advertising}
Web environment provides content publishers with a means to track user behavior in much greater detail than in an offline settings, including capturing user's registered information and activity logs of user's clicks, page views, searches, website visits, social activities, and interactions with ads. This allows for targeting of users based on their behavior, which is typically referred to as ad targeting \cite{ahmed2011scalable}. With the rise of big data applications and platforms, machine learning approaches are heavily leveraged to automate the ad targeting process. Within the past several years, there has been a plethora of research papers that explored different aspects of online advertising, each with the goal of maximizing the benefits for advertisers, content publishers, and users.

For any machine learning technique, features used to train a model typically have a critical influence on the performance of the deployed algorithm. Features derived from user events collected by publishers are often used in predicting the user's propensity to click or purchase \cite{djuric2014icdm}. However, these features represent only a weak proxy to what publishers and advertisers are actually interested in, namely, user's purchase intent. 
On the other hand, commercial e-mails in the form of promotions and purchase receipts convey a strong, very direct purchase intent signal that can enable advertisers to reach a high-quality audience. According to the Direct Marketing Association's ``National Client E-mail Report"\footnote{www.powerprodirect.com/statistics, accessed June 2015}, an e-mail user has been identified as having over $10\%$ more value than the average online customer. Despite these encouraging facts, limited work has been done to analyze the potential of features derived from commercial e-mails. 
A recent work \cite{grbovic2014generating} was among the first attempts to investigate the value of commercial e-mail data. The authors applied Sparse Principal Component Analysis (SPCA) on the counts of received e-mails in order to cluster commercial domains, demonstrating significant promise of e-mail data source.

Outside of the e-mail domain, information about user's purchase history is extensively used by e-commerce websites to recommend relevant products to their users \cite{Linden2003}. Recommendation systems predict which products a user will most likely be interested in either by exploiting purchase behavior of users with similar interests (referred to as collaborative filtering \cite{Linden2003}) or by using user's historical interaction with other products (i.e., context-based recommendation \cite{Zhang2013}). 
Unlike these studies, we are not limited to the data from a single website, as purchases extracted from e-mails enable us to gather information from hundreds of different e-commerce websites that can be exploited to learn better product predictions. To the best of our knowledge, this work represents the first study that offers a comprehensive empirical analysis and evaluation of product predictors using e-mail data of such scale and nature.

\subsection{Neural language models}
In a number of Natural Language Processing (NLP) applications, including information retrieval, part-of-speech tagging, chunking, and many others, specific objectives can all be generalized to the task of assigning a probability value to a sequence of words. To this end, language models have been developed, defining a mathematical model to capture statistical properties of words and the dependencies among them \cite{baeza1999modern,lavrenko2001relevance}. Traditionally, language model approaches represent each word as a feature vector using a one-hot representation, where a word vector has the same length as the size of a vocabulary, and the position that corresponds to the observed word is equal to 1, and 0 otherwise. However, this approach often exhibits significant limitations in practical tasks, suffering from high dimensionality of the problem and severe data sparsity, resulting in suboptimal performance.

Neural language models have been proposed to address these issues, inducing low-dimensional, distributed embeddings of words by means of neural networks \cite{bengio2006neural,collobert2011natural,turian2010word}. Such approaches take advantage of the word order in text documents, explicitly modeling the assumption that closer words in the word sequence are statistically more dependent. 
Historically, inefficient training of the neural network-based models has been an obstacle to their wider applicability, given that the vocabulary size may grow to several millions in practical tasks. However, this issue has been successfully addressed by recent advances in the field, particularly with the development of highly scalable continuous bag-of-words (CBOW) and skip-gram (SG) language models \cite{mikolov2013efficient,mikolov2013distributed} for learning word representations. These powerful, efficient models have shown very promising results in capturing both syntactic and semantic relationships between words in large-scale text corpora, obtaining state-of-the-art results on a plethora of NLP tasks. 

More recently, the concept of distributed representations has been extended beyond word representations to sentences and paragraphs \cite{djuric2015www,le2014distributed}, relational entities \cite{bordes2013translating,socher2013reasoning}, general text-based attributes \cite{kiros2014multiplicative}, descriptive text of images \cite{kiros2013multimodal}, nodes in graph structure \cite{perozzi2014deepwalk}, and other applications.

\begin{figure}[t]
\centering
{\includegraphics[width=0.25\textwidth]{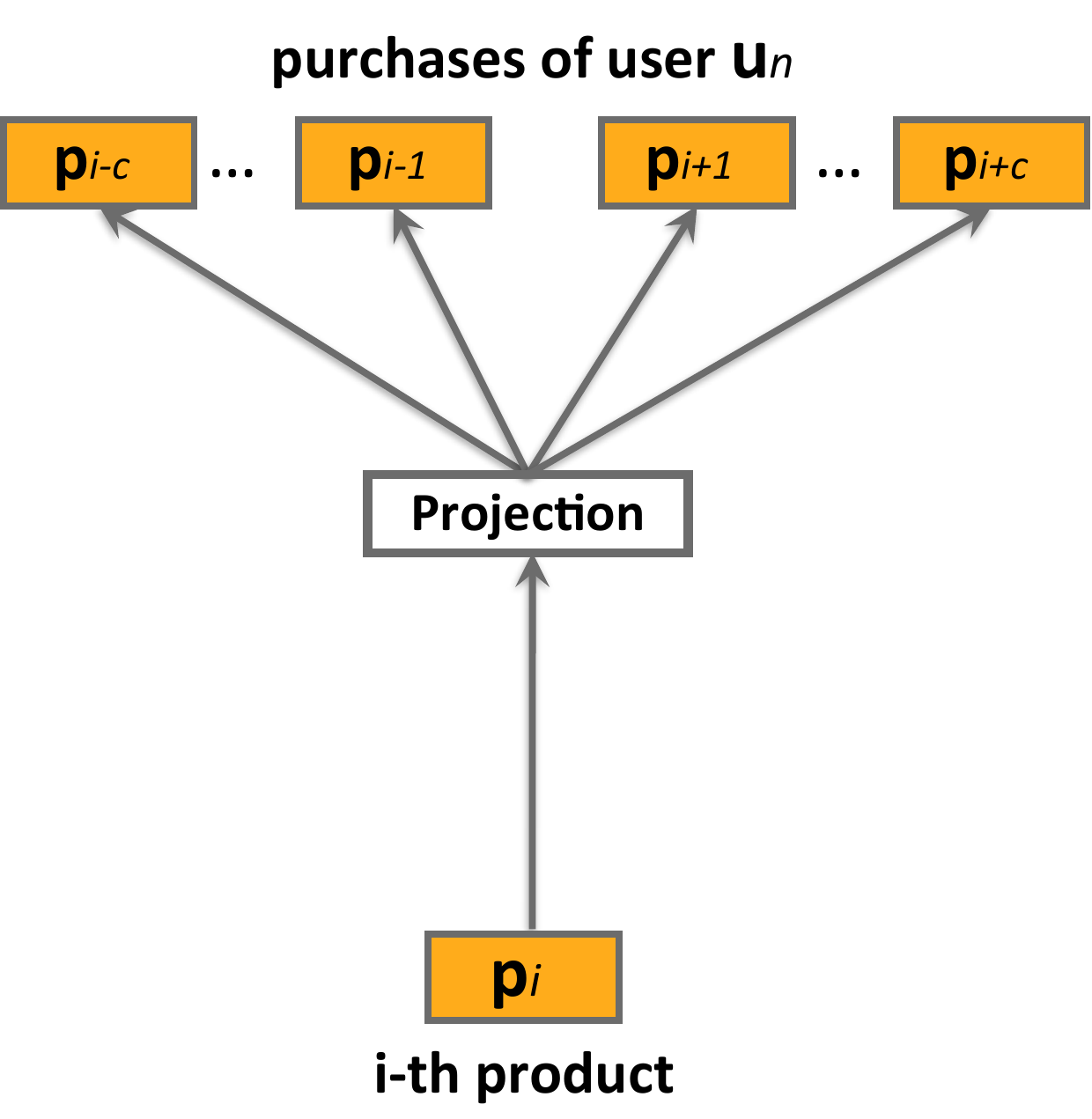}} 
\caption{prod2vec skip-gram model}
\label{fig:pur2vec}
\end{figure}

\section{Proposed approach}
\label{proposed}
In this section we describe the proposed methodology for the task of product recommendations, which leverages information about prior purchases determined from e-mail receipts. To address this task we propose to learn representation of products in low-dimensional space from historical logs using neural language models. Product recommendation can then be performed in the learned embedding space through simple nearest neighbor search. 

More specifically, given a set $\mathcal{S}$ of e-mail receipt logs obtained from $N$ users, where user's log $s = (e_{1}, \ldots, e_M) \in \mathcal{S}$ is defined as an uninterupted sequence of $M$ receipts, and each e-mail receipt $e_m = (p_{m1},p_{m2}, \ldots p_{mT_m})$ consists of $T_m$ purchased products, our objective is to find $D$-dimensional real-valued representation ${\bf v}_{p} \in \mathcal{R}^D$ of each product $p$ such that similar products lie nearby in the vector space. 

We propose several approaches for learning product representations that address specifics of the recommendations from e-mail receipts. We first propose \textit{prod2vec} method that considers all purchased products independently. We then propose novel \textit{bagged-prod2vec} method that takes into account that some products are listed as purchased together in e-mail receipts, which results in better, more useful product representations. Finally, we present product-to-product and user-to-product recommendation models that make use of the learned representations. 

\subsection{Low-dimensional product embeddings}

\textbf{prod2vec.} The prod2vec model involves learning vector representations of products from e-mail receipt logs by using a notion of a purchase sequence as a ``sentence'' and products within the sequence as ``words'', borrowing the terminology from the NLP domain (see Figure \ref{fig:pur2vec} for graphical representation of the model). More specifically, prod2vec learns product representations using the skip-gram model \cite{mikolov2013distributed} by maximizing the objective function over the entire set $\mathcal{S}$ of e-mail receipt logs, defined as follows
\begin{equation} \label{prod2vec_obj}
\mathcal{L} = \sum_{s \in \mathcal{S}} \sum_{p_i \in s} \sum_{-c\le j\le c, j\ne 0} \log \P(p_{i+j}|p_i),
\end{equation} 
where products from the same e-mail receipt are ordered arbitrarily. Probability $\P(p_{i+j}|p_i)$ of observing a neighboring product $p_{i+j}$ given the current product $p_i$ is defined using the soft-max function,
\begin{equation}\label{prod2vec_prq}
\P(p_{i+j}|p_i) = \frac{\exp(\mathbf{v}_{p_i}^\top \mathbf{v}_{p_{i+j}}^\prime)}{\sum_{p=1}^P \exp(\mathbf{v}_{p_i}^\top \mathbf{v}_{p}^\prime)},
\end{equation}
where $\mathbf{v}_{p}$ and $\mathbf{v}_{p}^\prime$ are the input and output vector representations of product $p$, $c$ is the length of the context for product sequences, and $P$ is the number of unique products in the vocabulary.
From equations \eqref{prod2vec_obj} and \eqref{prod2vec_prq} we see that prod2vec models context of product sequence, where products with similar contexts (i.e., with similar neighboring purchases) will have similar vector representations. However, prod2vec does not explicitly take into account that e-mail receipt may contain multiple products purchased at the same time, which we address by introducing a {\it bagged} version of prod2vec described below. 

\begin{figure}[t!]
\centering
\subfloat{\label{fig:tag_neighbors_a}\includegraphics[width=0.21\textwidth]{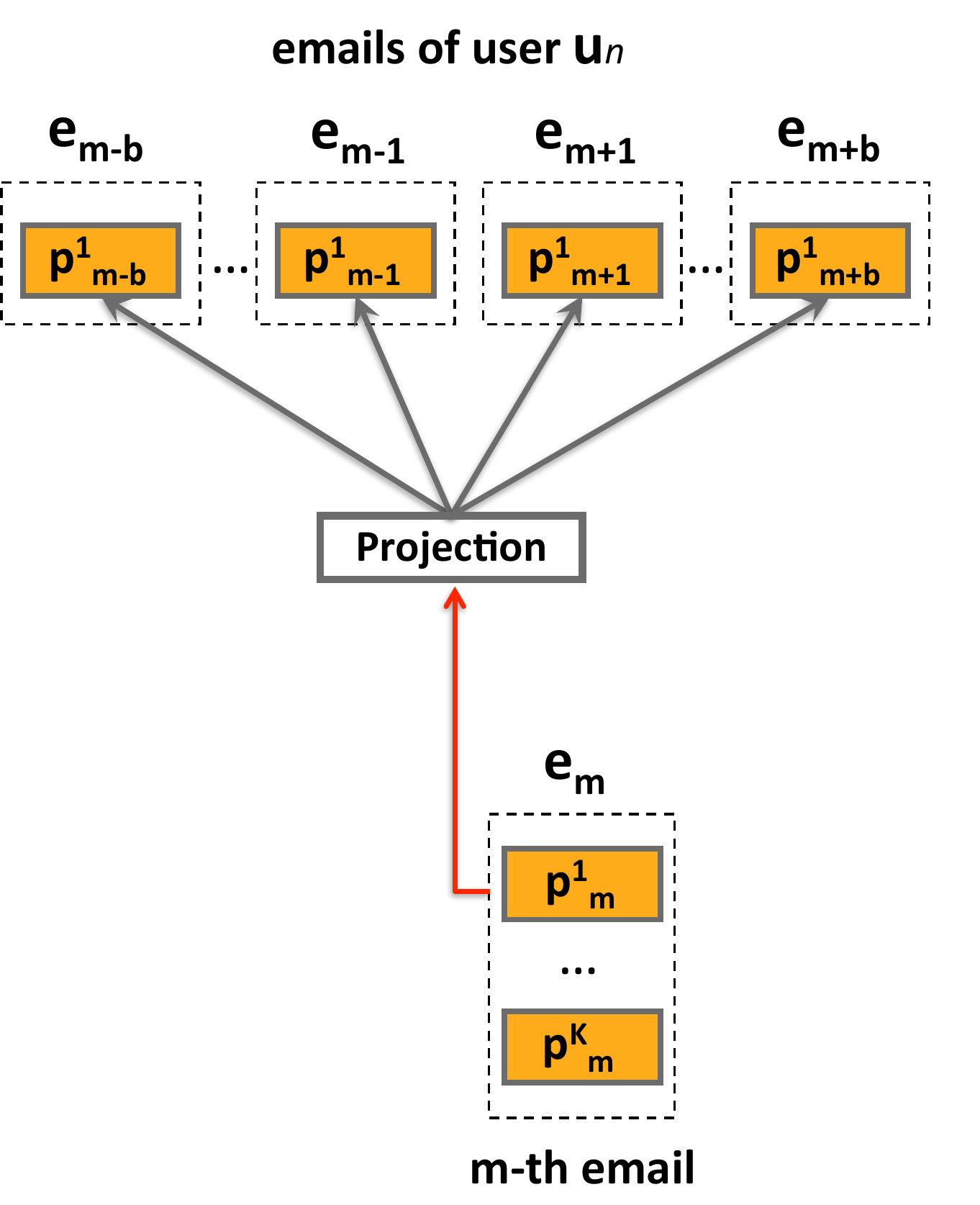}}
\subfloat{\label{fig:tag_neighbors_b}\includegraphics[width=0.21\textwidth]{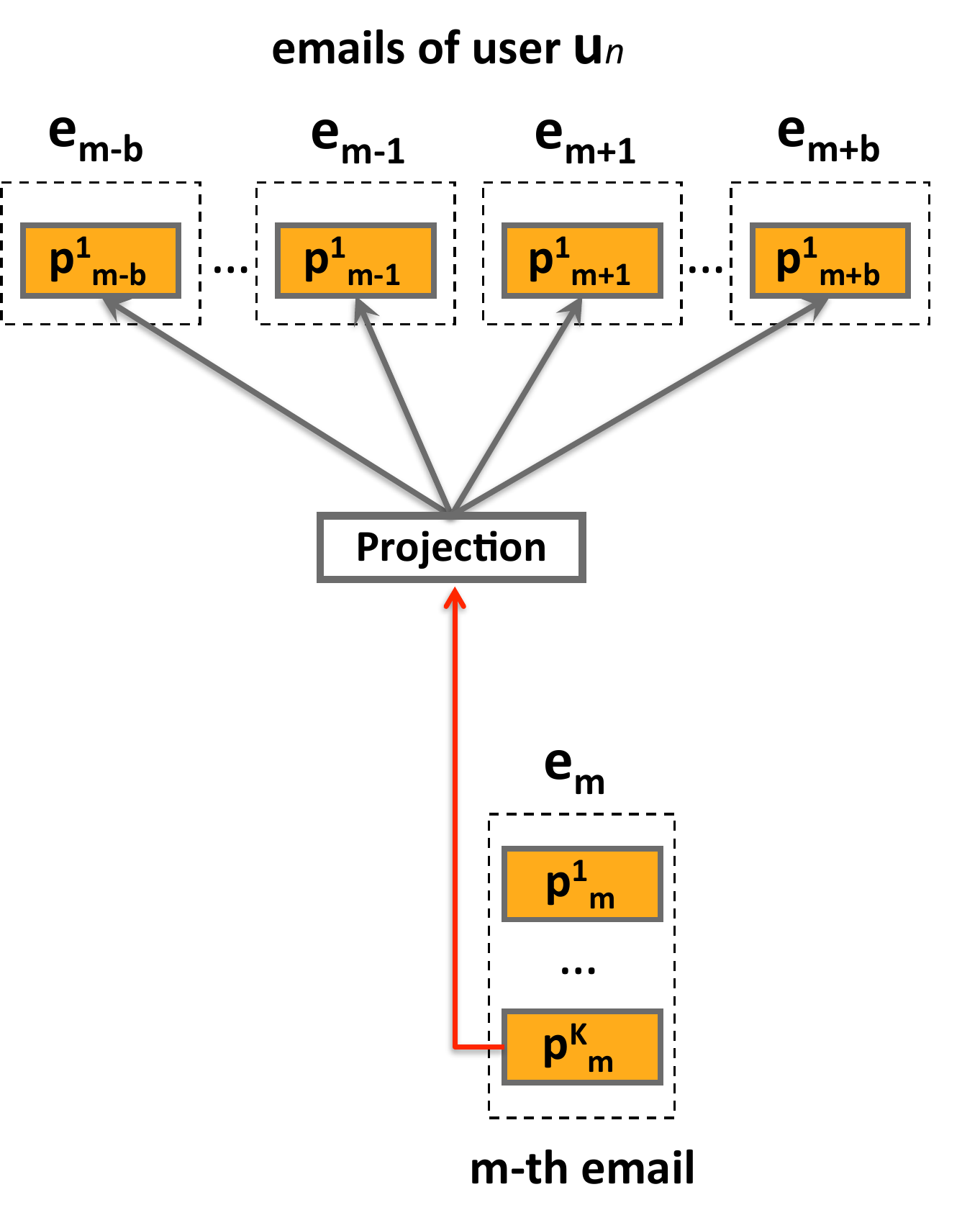}} 
\caption{bagged-prod2vec model updates} 
\label{fig:skip_bag}
\end{figure}

\textbf{bagged-prod2vec.} In order to account for the fact that multiple products may be purchased at the same time, we propose a modified skip-gram model that introduces a notion of a shopping bag. As depicted in Figure~\ref{fig:skip_bag}, the model operates at the level of e-mail receipts instead at the level of products. Product vector representations are learned by maximizing a modified objective function over e-mail sequences $s$, defined as follows
\begin{equation} \label{email2vec_obj}
\mathcal{L} = \sum_{s \in \mathcal{S}} \sum_{e_m \in s} \sum_{-n\le j\le n, j\ne 0} \sum_{k = 1, \ldots, T_{m}} \log \P(e_{m+j}|p_{mk}).
\end{equation}
Probability $\P(e_{m+j}|p_{mk})$ of observing products from neighboring e-mail receipt $e_{m+j}$, $e_{m+j} = (p_{m+j,1} \ldots p_{m+j,T_{m}})$, given the $k$-th product from $m$-th e-mail receipt reduces to a product of probabilities $\P(e_{m+j}|p_{mk}) =  \P(p_{m+j,1}|p_{mk}) \times \ldots \times \P(p_{m+j,T_{m}}|p_{mk}) $, each defined using soft-max \eqref{prod2vec_prq}. 
Note that the third sum in \eqref{email2vec_obj} goes over receipts, so the items from the same e-mail receipt do not predict each other during training. In addition, in order to capture temporal aspects of product purchases we propose to use the directed language model, where as context we only use future products \cite{grbovic2015www}. The modification allows us to learn product embeddings capable of predicting future purchases.

\textbf{Learning.} The models were optimized using stochastic gradient ascent, suitable for large-scale problems. However, computation of gradients $\nabla \mathcal{L}$ in \eqref{prod2vec_obj} and \eqref{email2vec_obj} are proportional to the vocabulary size $P$, which is computationally expensive in practical tasks as $P$ could easily reach millions of products. As an alternative, we used negative sampling approach proposed in \cite{mikolov2013distributed}, which significantly reduces the computational complexity.

\subsection{Product-to-product predictive models}
\label{i2i}
Having learned low-dimensional product representations, we considered several possibilities for predicting the next product to be purchased. 

\textbf{prod2vec-topK.} Given a purchased product, the method calculates cosine similarities to all other products in the vocabulary and recommends the top $K$ most similar products. 

\textbf{prod2vec-cluster.} To be able to make more diverse recommendations, we considered grouping similar products into clusters and recommending products from a cluster that is related  to the cluster of previously purchased product. 
We applied $K$-means clustering algorithm implemented on the top of Hadoop distributed system where we grouped products based on cosine similarity between their low-dimensional representations. We assume that purchasing a product from any of the $C$ clusters after a purchase from cluster $c_i$ follows a multinomial distribution $Mu(\theta_{i1},\theta_{i2},\ldots \theta_{iC})$, where $\theta_{ij}$ is the probability that a purchase from cluster $c_i$ is followed by a purchase from cluster $c_j$. In order to estimate parameters $\theta_{ij}$, for each $i$ and $j$, we adopted a maximum likelihood approach,
\begin{equation} \label{mle_theta}
\hat{\theta}_{ij} = \frac{\text{\# of times $c_i$ purchase was followed by $c_j$}}{\text{count of $c_i$ purchases}}.
\end{equation}

In order to recommend a new product given a purchased product $p$, we first identify which cluster $p$ belongs to (e.g., $p \in c_i$). Next, we rank all clusters $c_j$, $j=1, \ldots, C$, by the value of $\theta_{ij}$ and consider the top ones as top-related clusters to cluster $c_i$. Finally, products from the top clusters are sorted by their cosine similarity to $p$, and we use the top $K$ products as recommendations.

\subsection{User-to-product predictive models}
\label{u2i}

\begin{figure}[t!]
\centering
\subfloat{\includegraphics[width=0.35\textwidth]{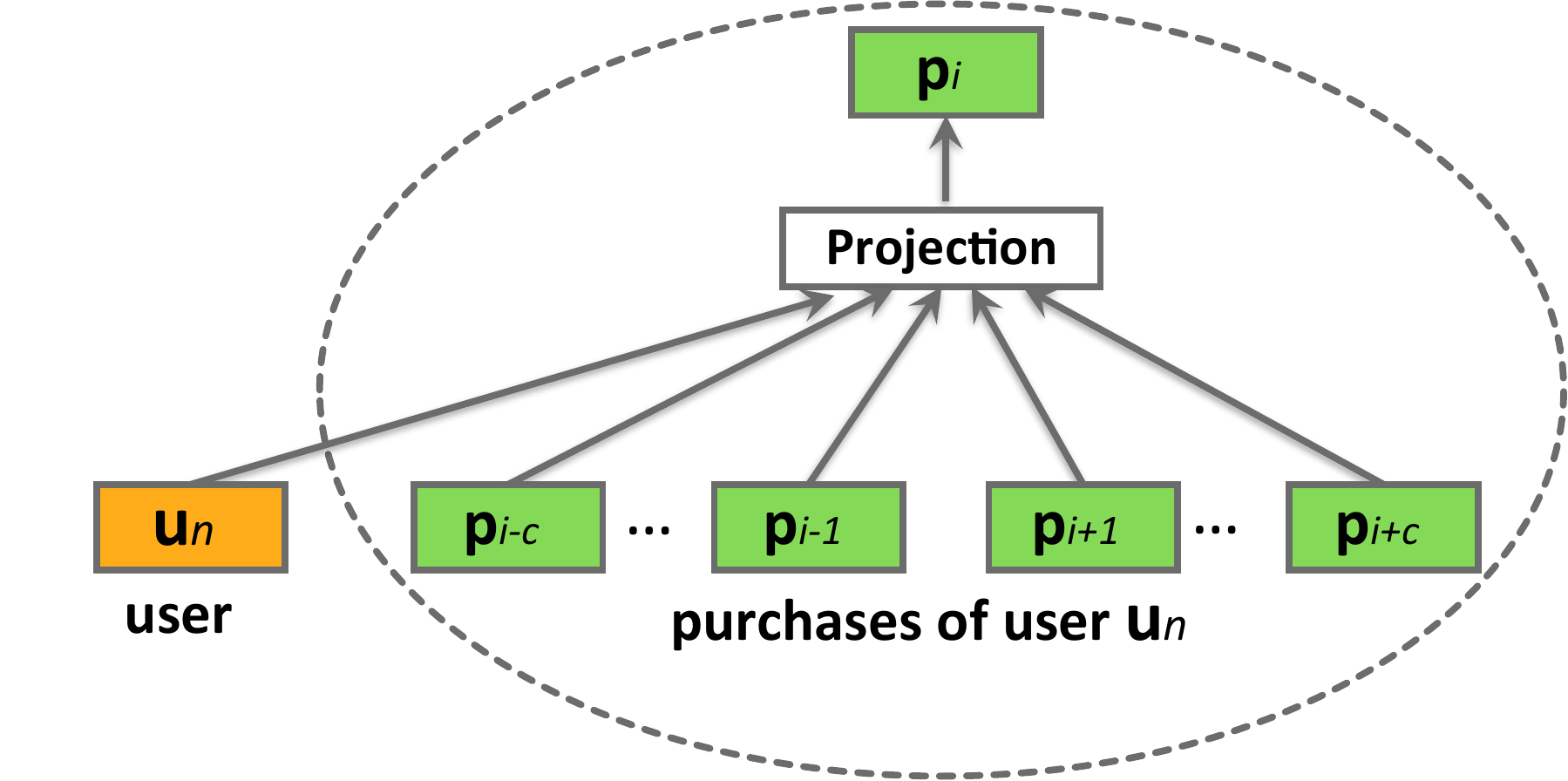}}
\caption{User embeddings for user to product predictions} 
\label{fig:user2vec model}
\end{figure}

In addition to product-to-product predictions \cite{katukuri2014recommending,katukuri2013large}, most recommendation engines allow user-to-product predictions as well \cite{hu2014style,zhao2014we}. Recommendation for a user are typically made considering historical purchases and/or interests inferred using other data sources such as user's online behavior \cite{hu2014style}, social contacts \cite{zhao2014we}, and others. In this section we propose a novel approach to simultaneously learn vector representations of products and users such that, given a user, recommendations can be performed by finding $K$ nearest products in the joint embedding space. 

\textbf{user2vec.} The user2vec model simultaneously learns vector representations of products and users by considering the user as a ``global context'', motivated by paragraph2vec algorithm \cite{le2014distributed}. The architecture of such model is illustrated in Figure~\ref{fig:user2vec model}. The training data set was derived from user purchase sequences $\mathcal{S}$, which comprised users $u_n$ and their purchased products ordered by the time of purchase, $u_n=(p_{n1},p_{n2}, \ldots p_{nU_n})$, where $U_n$ denotes number of items purchased by user $u_n$. During training, user vectors are updated to predict the products from their e-mail receipts, while product vectors are learned to predict other products in their context. For simplicity of presentation and w.l.o.g. in the following we present non-bagged version of the language model, however we note that it is straightforward to extend the presented methodology to use the bagged version. 

More specifically, objective of user2vec is to maximize the log-likelihood over the set $\mathcal{S}$ of all purchase sequences,
\begin{equation} \label{user2vec_obj}
\begin{aligned}
\mathcal{L} = &  \sum_{s \in \mathcal{S}} \Big( \sum_{u_n \in s}  \log \P(u_n|p_{n1}:p_{nU_n}) \\
& + \sum_{p_{ni}\in u_n} \log \P(p_{ni}|p_{n,i-c}:p_{n,i+c}, u_n) \Big)
\end{aligned}
\end{equation}
where $c$ is the length of the context for products in purchase sequence of the $n$-th user. The probability $\P(p_{ni}|p_{n,i-c}:p_{n,i+c},u_n)$ is defined using a soft-max function,
\begin{equation} \label{user2vec_prw}
\P(p_{ni} |p_{n,i-c}:p_{n,i+c},u_n) = \frac{\exp(\bar{\mathbf{v}}^\top\mathbf{v}_{p_{ni}}^\prime)}{\sum_{p=1}^V\exp(\bar{\mathbf{v}}^\top\mathbf{v}_{p}^\prime)},
\end{equation}
where $\mathbf{v}_{p_{ni}}^\prime$ is the output vector representation of $p_{ni}$, and $\bar{\mathbf{v}}$ is averaged vector representation of the product context including corresponding $u_n$, defined as
\begin{equation} \label{user2vec_avgw}
\bar{\mathbf{v}}= \frac{1}{2c + 1}(\mathbf{v}_{u_n} + \sum_{-c\le j\le c,j\ne 0} \mathbf{v}_{p_{n,i+j}}),
\end{equation}
where $\mathbf{v}_{p}$ is the input vector representation of $p$. Similarly, the probability $\P(u_n|p_{n1}:p_{nU_n})$ is defined as
\begin{equation} \label{user2vec_prq}
\P(u_n |p_{n1}:p_{nU_n}) = \frac{\exp({\bar{\mathbf{v}}_n}^\top\mathbf{v}_{u_{n}}^\prime)}{\sum_{p=1}^V\exp({\bar{\mathbf{v}}_n}^\top\mathbf{v}_{p}^\prime)},
\end{equation}
where $\mathbf{v}_{u_n}^\prime$ is the output vector representation of $u_n$, and $\bar{\mathbf{v}}_n$ is averaged input vector representation of all the products purchased by user $u_n$,
\begin{equation} \label{user2vec_avgq}
\bar{\mathbf{v}}_n= \frac{1}{U_n} \sum_{i=1}^{U_n} \mathbf{v}_{p_{ni}}.
\end{equation}

One of the main advantages of the {\it user2vec} model is that the product recommendations are specifically tailored for that user based on his purchase history. However, disadvantage is that the model would need to be updated very frequently. Unlike product-to-product recommendations, which may be relevant for longer time periods, user-to-product recommendations need to change often to account for the most recent user purchases.

\section{Experiments}
The experimental section is organized as follows. We first describe data set used in development of our product-to-product and user-to-product predictors. Next we present valuable insights regarding purchase behavior of different age and gender groups in different US states. This information can be leveraged to improve demo- and geo-targeting of user groups. This is followed by a section on effectiveness of recommendation of popular products in different user cohorts, including age, gender, and US state. Finally, we show comparative results of various baseline recommendation algorithms. We conclude with the description of the system implementation at a large scale and bucket results that preceded our product launch.

\subsection{Data set}

Our data sets included e-mail receipts sent to users who voluntarily opted-in for such studies, where we anonymized user IDs. Message bodies were analyzed by automated systems. Product names and prices were extracted from e-mail messages using an in-house extraction tool.

Training data set collected for purposes of developing product prediction models comprised of more than $280.7$ million purchases made by $N=29$ million users from $172$ commercial websites. The product vocabulary included $P=2.1$ million most frequently purchased products priced over $\$5$.

More formally, data set $\mathcal{D}_{p} = \{(u_n, s_n),  n=1,...,N\}$ was derived by forming e-mail receipt sequences $s_n$ for each user  $u_n$, along with their timestamps. Specifically, $s_n=\{({e}_{n1}, t_{n1}), \ldots, ({e}_{nM_n}, t_{nM_n})\}$, where ${ e}_{nm}$ is a receipt comprising one or more purchased products, and $t_{nm}$ is receipt timestamp.
Predictions were evaluated on a held-out month of user purchases $D_{p}^{ts}$, formed in the same manner as $D_{p}$. We measured the effectiveness of recommendations using prediction accuracy. In particular, we measured the number of product purchases that were correctly predicted, divided by the total number of purchases.
For all models, the accuracy was measured separately on each day, based on the recommendations calculated using prior days. We set a daily budget of distinct recommendations each algorithm is allowed to make for a user to $K=20$, based on an estimate of optimal number of ads users can effectively perceive daily \cite{cho2004people}. 

\subsection{Insights from purchase data}

\begin{figure}[t]
\centering
\subfloat[Percentage of purchasing users among all online users]{\includegraphics[width = 1.64in]{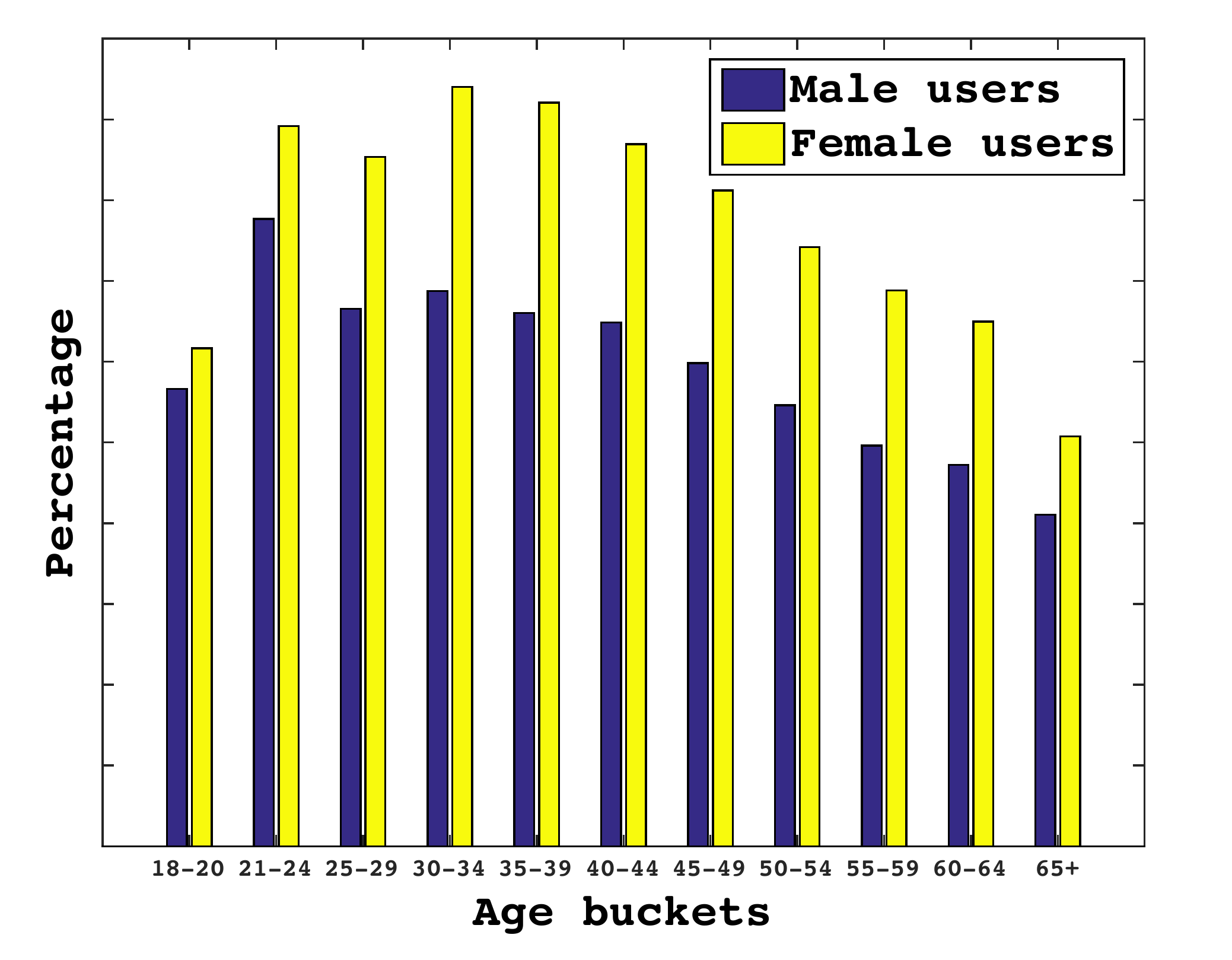}}
\subfloat[Average product price]{\includegraphics[width = 1.64in]{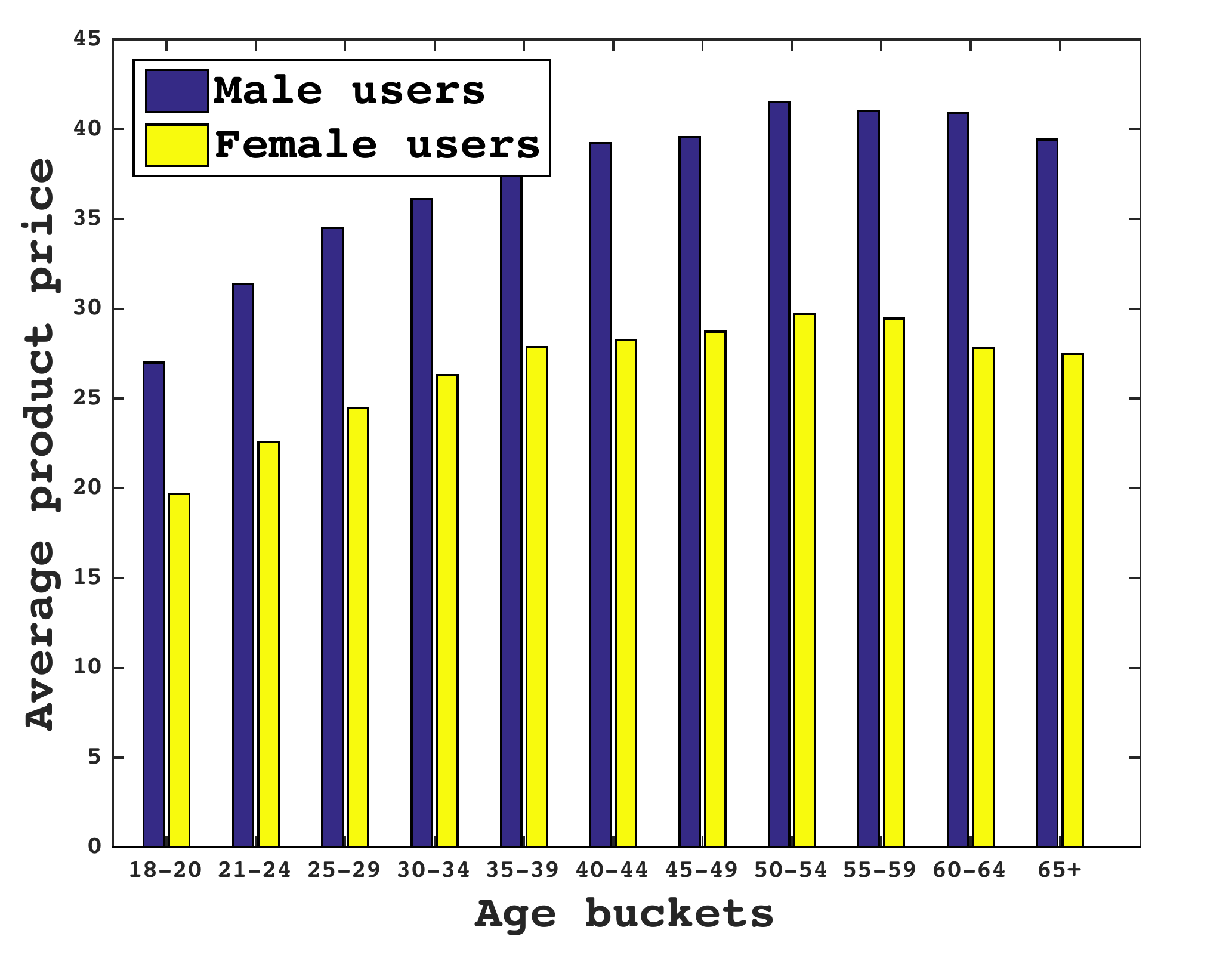}} \\
\caption{Purchasing habits for different demographics}
\label{fig:demo_purch}
\end{figure}

\begin{figure*}[t]
\subfloat[ages 18-20]{\includegraphics[width = 1.7in]{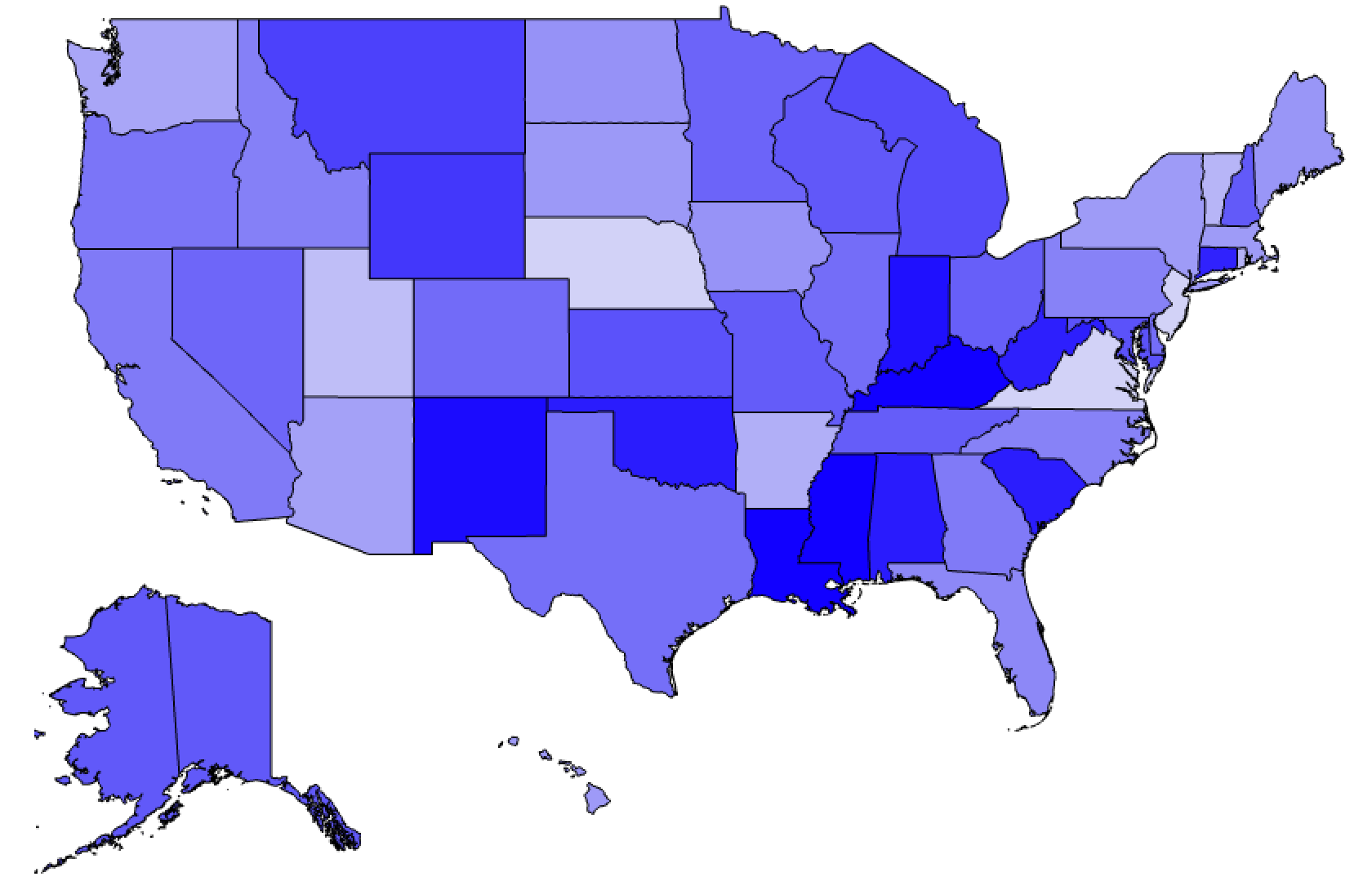}}
\subfloat[ages 40-44]{\includegraphics[width = 1.7in]{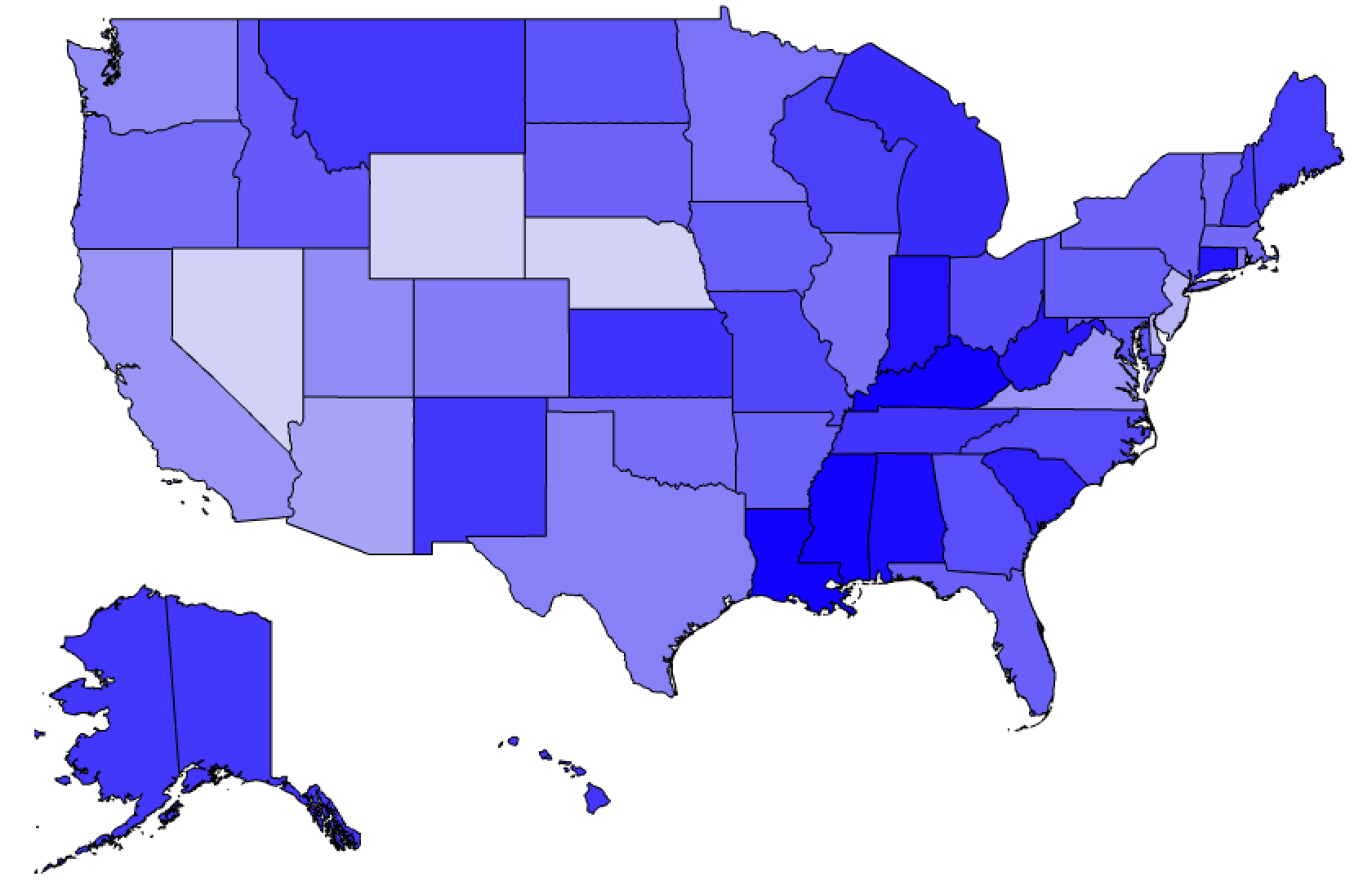}}
\subfloat[ages 18-20]{\includegraphics[width = 1.7in]{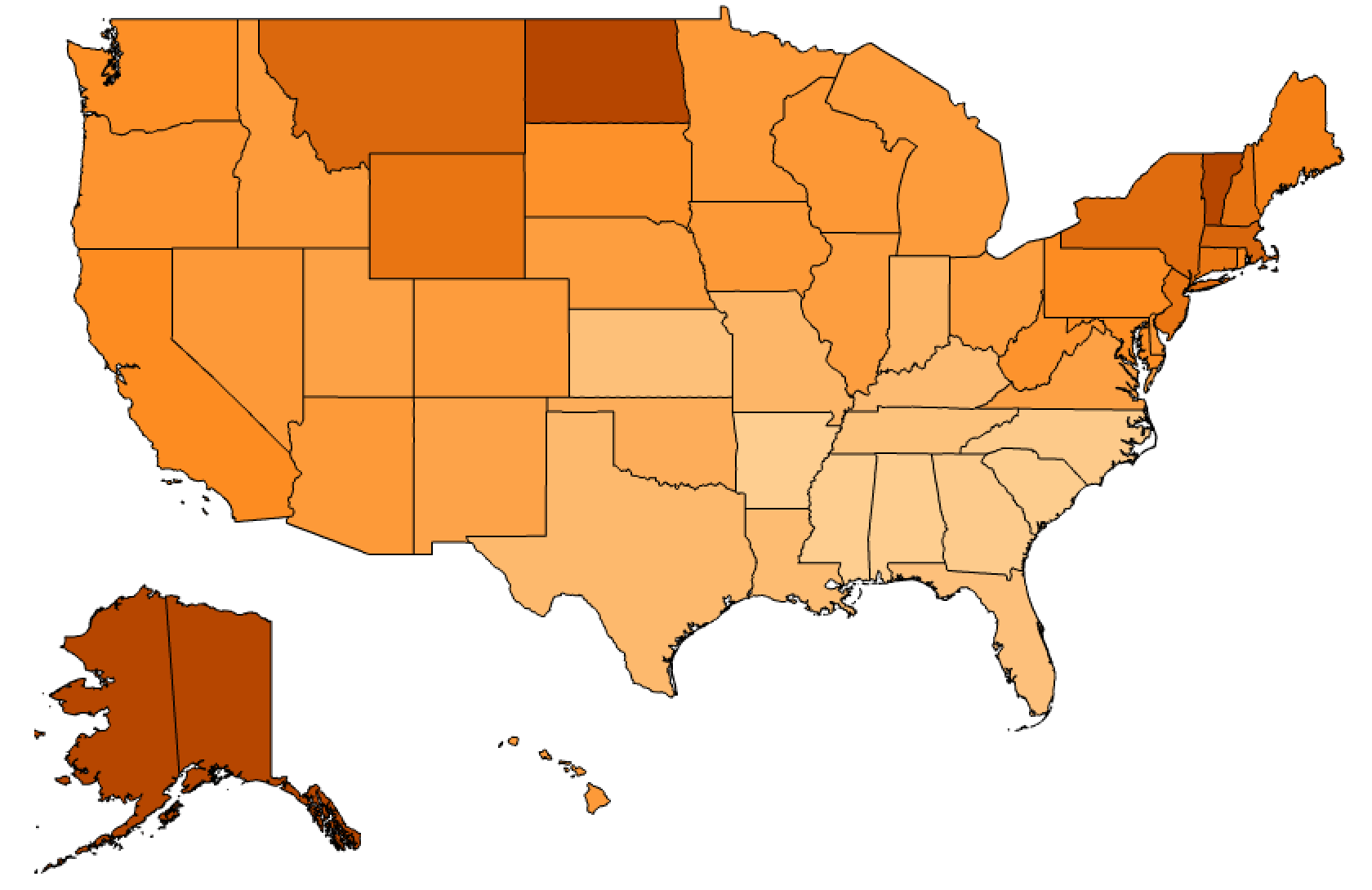}} 
\subfloat[ages 40-44]{\includegraphics[width = 1.7in]{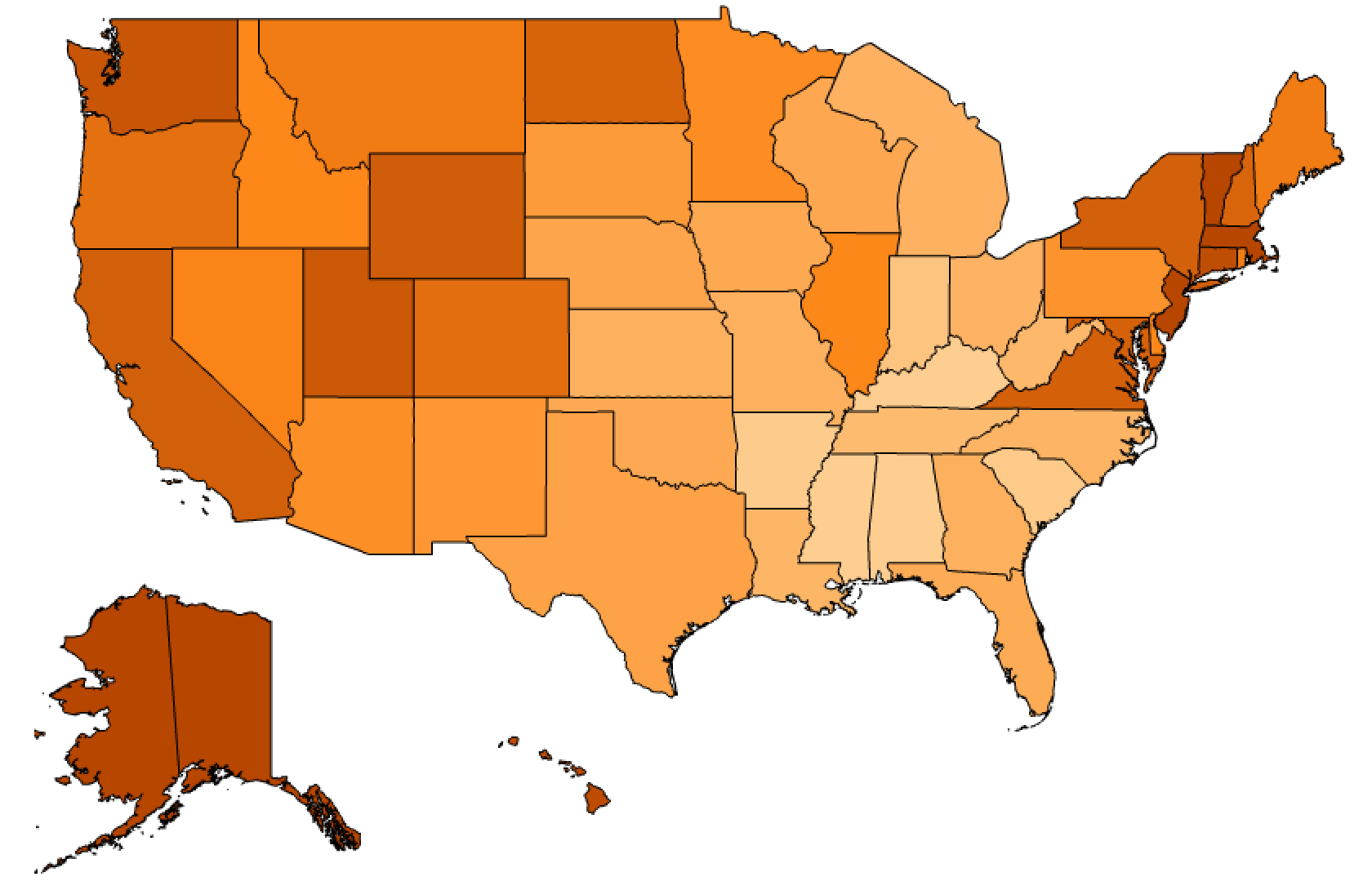}}\\
\subfloat[ages 18-20]{\includegraphics[width = 1.7in]{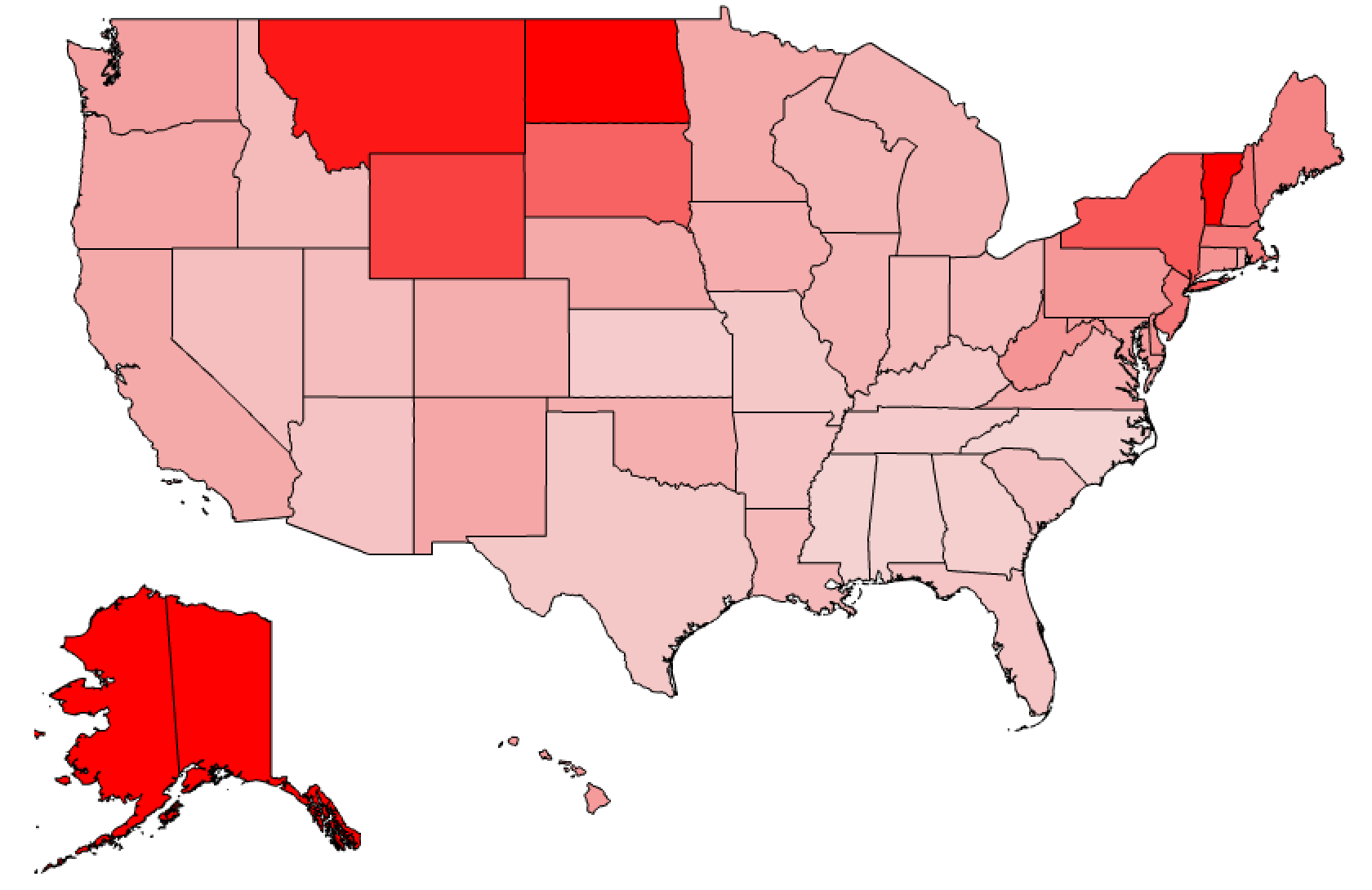}} 
\subfloat[ages 40-44]{\includegraphics[width = 1.7in]{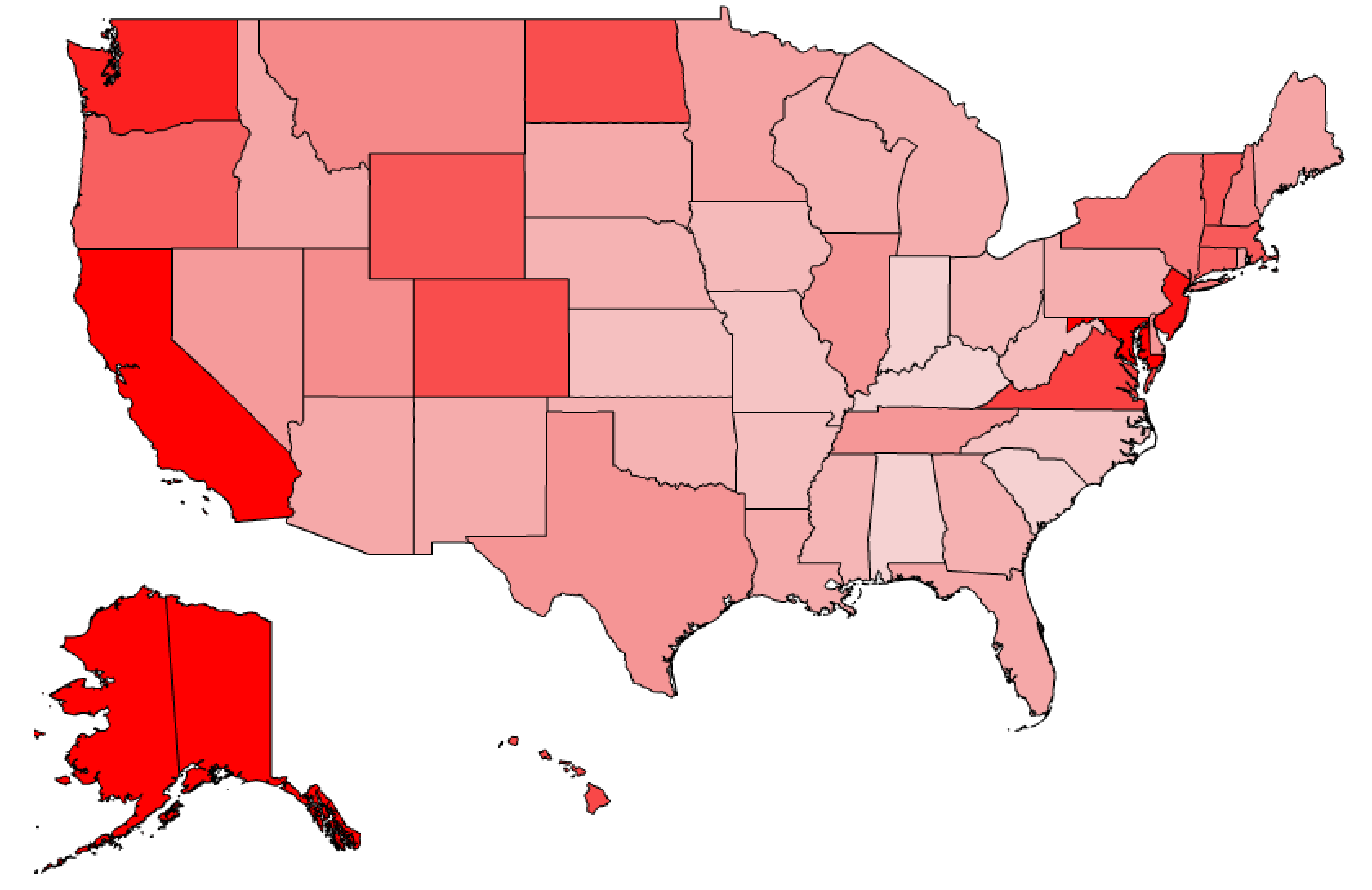}}
\subfloat[ages 18-20]{\includegraphics[width = 1.7in]{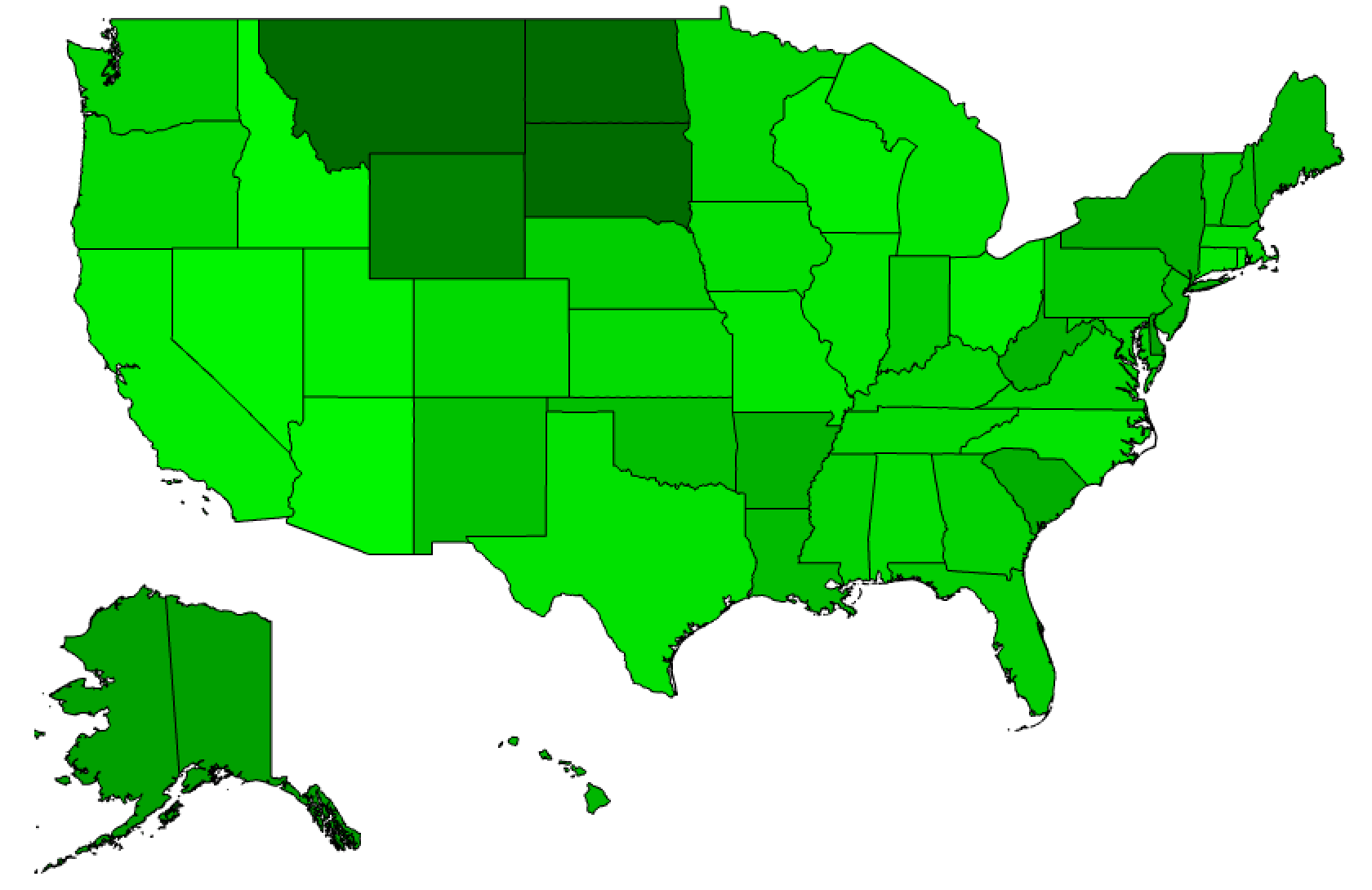}} 
\subfloat[ages 40-44]{\includegraphics[width = 1.7in]{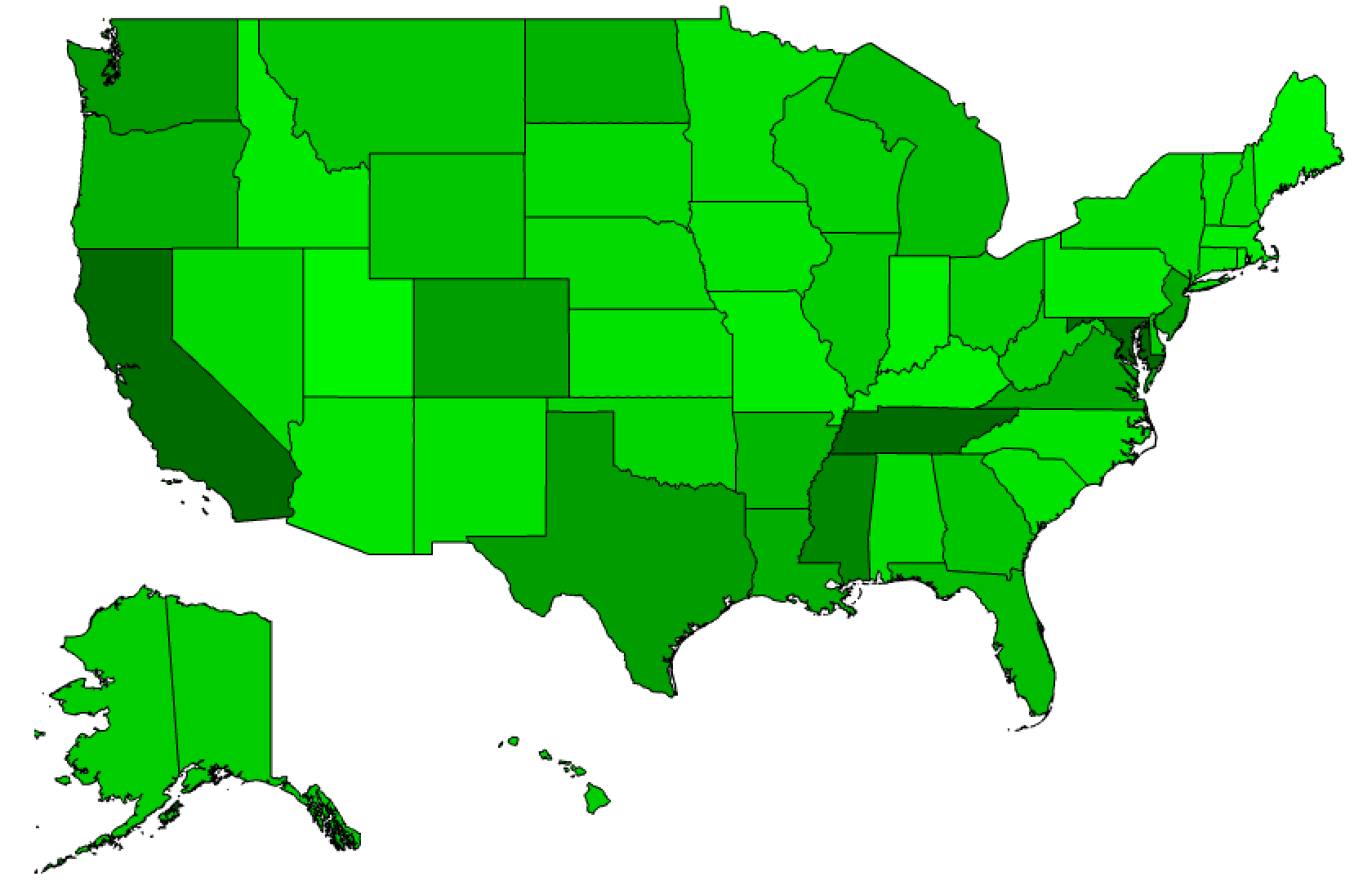}}\\
\caption{Purchasing behavior for different cohorts, dark color encodes higher values: (a, b) Percentage of shoppers among online users; (c, d) Average number of purchases per user; (e, f) Average amount spent per user; (g, h) Average product price}
\label{fig_states}
\end{figure*}

To get deeper insights into behavior of online users based on their demographic background and geographic location, we segregated users into cohorts based on their age, gender, and location, and looked at their purchasing habits. Such information can be very valuable to marketers when considering how and where to spend their campaign budget. We only considered users from the US, and computed statistics per each state separately. First, we were interested in differences between male and female shopping behavior for different age ranges. In addition, we were interested in the following aspects of purchasing behavior: 1) percentage of online users who shop online; 2) average number of products bought; 4) average spent per user; and 4) average price of a bought item. Figures \ref{fig:demo_purch} and \ref{fig_states} illustrate the results.

In Figure \ref{fig:demo_purch} for each gender and age group we show the percentage of online users who shop online, as well as the average prices of bought items. Expectedly, we see that male and female demographics exhibit different shopping patterns. Throughout the different age buckets percentage of female shoppers is consistently higher than for males, peaking in the 30-34 age range. On the other hand, percentage of male shoppers reaches its maximum in the 21-24 bucket, and from there drops steadily. In addition, we observe that male users buy more expensive items on average.

Furthermore, in Figure \ref{fig_states} we show per-state results, where darker color indicates higher values. Note that we only show results for different locations and age ranges, as we did not see significant differences between male and female purchasing behavior across states. First, in Figures \ref{fig_states}(a) and \ref{fig_states}(b) we show the percent of online shoppers in each state, for younger (18-20) and medium-age (40-44) populations, where we can see there exist significant differences between the states. In Figures \ref{fig_states}(c) and \ref{fig_states}(d) we show the average number of purchased products per state. Here, for both age buckets we see that the southern states purchase the least number of items. However, there is a significant difference between younger and older populations, as in northern states younger populations purchase more products, while in the northeast and the western states this holds true for the older populations. 
If we take a look at Figures \ref{fig_states}(e) and \ref{fig_states}(f) where we compare the average amount of money spent per capita, we can observe similar patterns, where in states like Alaska and North Dakota younger populations spend more money than peers from other states, and for older populations states with highest spending per user are California, Washington, and again Alaska. Interestingly, users from Alaska are the biggest spenders, irrespective of the age or metric used. Similar holds when we consider average price of a purchased item, shown in Figures \ref{fig_states}(g) and \ref{fig_states}(h).

\subsection{Recommending popular products}
In this section we evaluate predictive properties of popular products. Recommending popular products to global population is a common baseline due to its strong performance, especially during holidays (e.g., Christmas, Thanksgiving). Such recommendations are intuitive, easy to calculate and implement, and may serve in a cold start scenarios for newly registered users when we have limited information.

\begin{figure}[t]
\centering
{\includegraphics[width=0.40\textwidth]{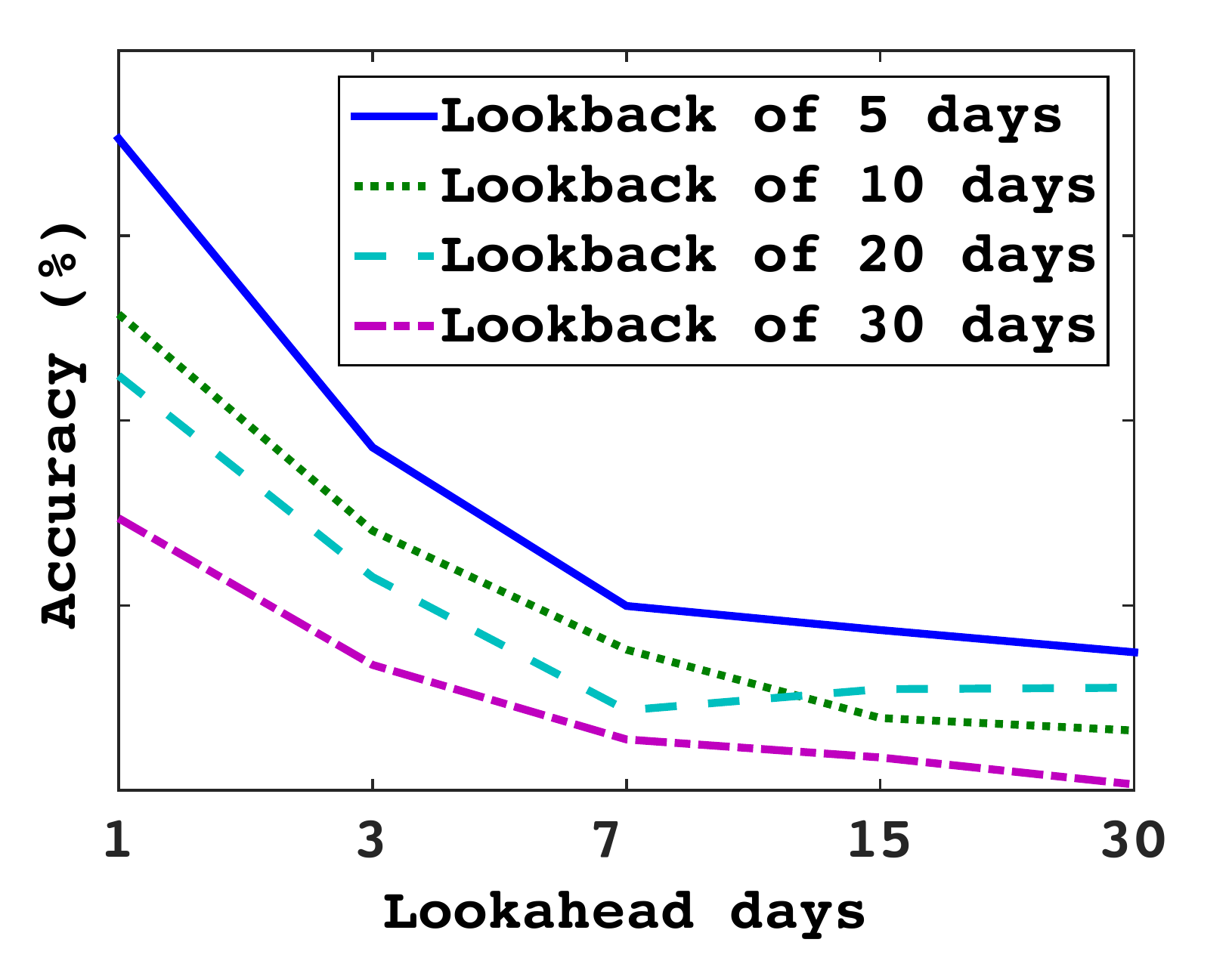}} 
\caption{Prediction accuracy and longevity of popular products with different lookbacks}
\label{fig:global}
\end{figure}

We considered how far back we need to look when calculating popularity and how long popular products stay relevant. In Figure~\ref{fig:global} we give results for popular products calculated in the previous $5, 10, 20$, and $30$ days of training data $D_p$, evaluated on the first $1, 3, 7, 15$, and $30$ days of test data $D_{p}^{ts}$. To account for ever-changing user purchase tastes, the results indicate that popular products need to be recalculated at least every $3$ days with lookback of at most $5$ days.

\begin{figure}[t]
\centering
{\includegraphics[width=0.40\textwidth]{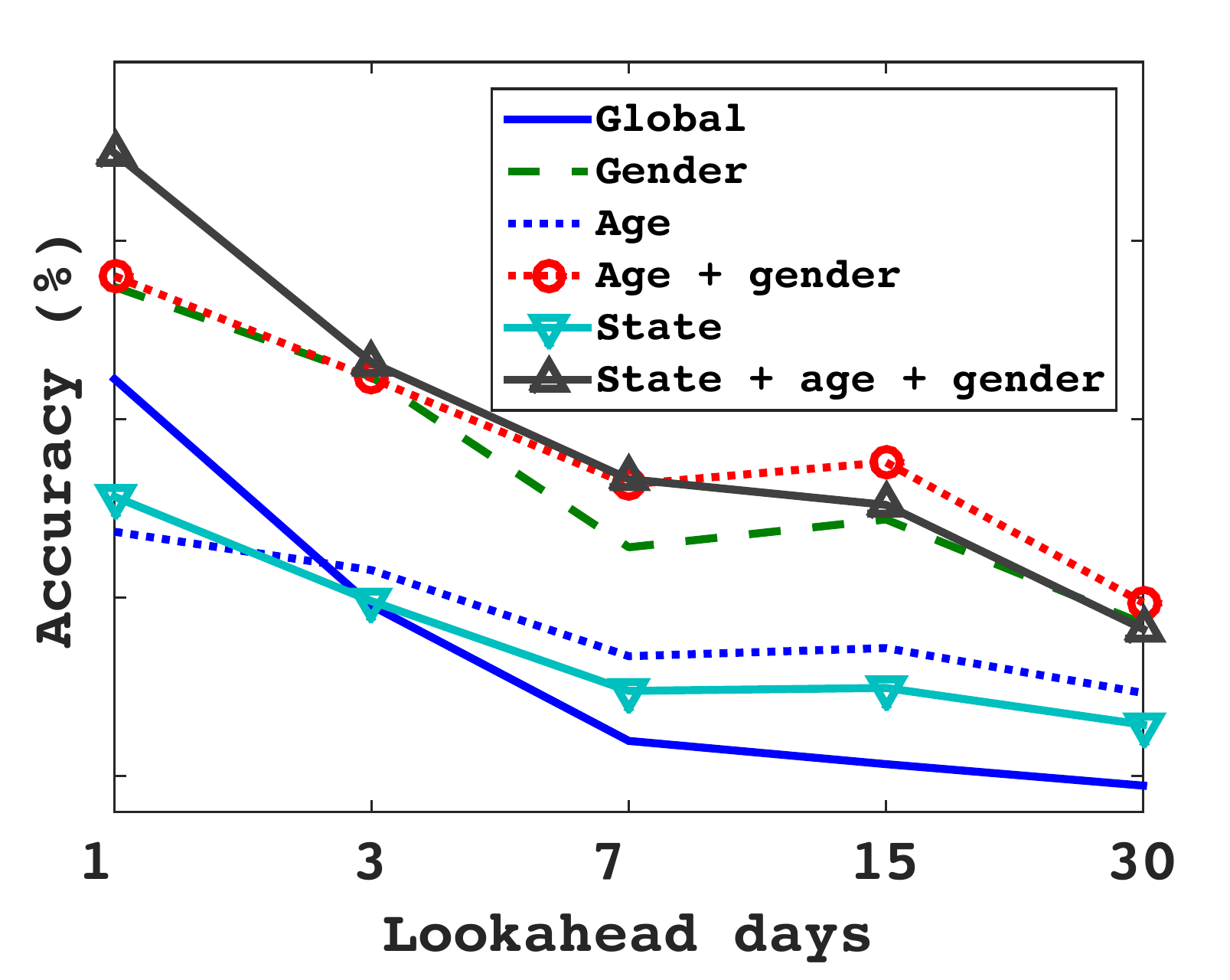}} 
\caption{Prediction accuracy of popular products for different user cohorts}
\label{fig:global_cohort}
\end{figure}

Next, we evaluated the prediction accuracy of popular products computed for different user cohorts, and compared to the accuracy of globally popular products. In Figure~\ref{fig:global_cohort} we compare accuracies of popular products in different user cohorts, with the lookback fixed at $5$ days. We can observe that popular products in gender cohorts give bigger lift in prediction accuracy than popular products in age or state cohorts. Further, jointly considering age and gender when calculating popular products outperforms the popular gender products. Finally, including geographic dimension contributes to further accuracy lift.
Overall, the results indicate that in the cold start scenario the best choice of which popular products to recommend are the ones from the user's (age, gender, location) cohort. The results also suggest that popular products should be recalculated often to avoid decrease in accuracy with every passing day.

\begin{table*}[t]
\centering
{\scriptsize
\caption{Examples of product recommendations made by the bagged-prod2vec model}
\begin{tabular}{ l l l }
{\bf despicable me } & {\bf first aid for the usmle step 1} & {\bf disney frozen lunch napkins } \\
\hline \hline
\rowcolor{lightgray}
monsters university  & usmle step 1 secrets 3e & disneys frozen party 9 square lunchdinner plates  \\
the croods & first aid basic sciences 2e & disneys frozen party 9oz hotcold cups \\
\rowcolor{lightgray}
turbo &  usmle step 1 qbook & disneys frozen party 7x7 square cakedessert plates \\
cloudy with a chance of meatballs & brs physiology  & disneys frozen party printed plastic tablecover  \\
\rowcolor{lightgray}
hotel transylvania  &rapid review pathology with student consult &  disneys frozen party 7 square cakedessert plates   \\
brave & lippincotts microcards microbiology flash cards  & disney frozen beverage napkins birthday party supplies \\
\rowcolor{lightgray}
the smurfs & first aid cases for the usmle step 2 & disney frozen 9 oz paper cups\\
wreckit ralph & highyield neuroanatomy & frozen invitation and thank you card \\
\rowcolor{lightgray}
planes & lange pharmacology flash cards third edition & disneys frozen party treat bags  \\
\bottomrule
\end{tabular}
\label{tab:prod2vec_pred}
}
\end{table*}

\begin{table*}[t]
\centering
{\scriptsize
\caption{Product recommendations for product {\it cressi supernova dry snorkel} }
\begin{tabular}{ l | l l}

{\bf bagged-prod2vec-topK } & {\bf bagged-prod2vec-cluster}  & {\bf cluster ID}  \\
\hline \hline
jaws quick spit antifog 1 ounce & cressi neoprene mask strap   \\
cressi neoprene mask strap & cressi frameless mask & cluster 1 \\
cressi frameless mask &  cressi scuba diving snorkeling freediving mask snorkel set \\
\cline{2-3}
akona 2 mm neoprene low cut socks &  akona 2 mm neoprene low cut socks \\
tilos neoprene fin socks &  tilos neoprene fin socks  & cluster 2  \\
cressi scuba diving snorkeling freediving mask snorkel set & mares equator 2mm dive boots   \\ \cline{2-3}
mares cruise mesh duffle bag & jaws quick spit antifog 1 ounce  \\
us divers island dry snorkel & aqua sphere kayenne goggle with clear lens clear black regular & cluster 3 \\
us divers trek travel fin & aqua sphere kayenne goggle with clear lens black regular   \\
\cline{2-3}
us divers proflex ii diving fins & nikon coolpix aw120 161 mp wi-fi and waterproof digital camera \\
mares cruise backpack mesh bag & olympus stylus tg ihs digital camera with 5x optical zoom & cluster 4 \\
water gear fin socks & nikon coolpix aw110 wi fi and waterproof digital camera with gps \\
\bottomrule
\end{tabular}
\label{tab:doc2vec_pred1}
}
\end{table*}

\subsection{Recommending predicted products}
\label{sec:predicted}

In this section we experiment with recommending products to users based on neural language models described in Section~\ref{proposed}. Specifically, we compare the following algorithms:

{\bf 1) $\text{prod2vec-topK}$} was trained using data set $D_{p}$, where product vectors were learned by maximizing log-likelihood of observing other products from sequences $s$, as proposed in \eqref{prod2vec_obj}. Recommendations for a given product $p_i$ were given by selecting the top $K$ most similar products based on the cosine similarity in the resulting vector space.

{\bf 2) $\text{bagged-prod2vec-topK}$} was trained using $D_{p}$, where product vectors were learned by maximizing log-likelihood of observing other products from e-mail sequences $s$ as proposed in \eqref{email2vec_obj}. Recommendations for a given product $p_i$ were given by selecting the top $K$ most similar products based on the cosine similarity in the resulting vector space.

{\bf 3) $\text{bagged-prod2vec}$-cluster} was trained similarly to the \text{bagged-prod2vec} model, followed by clustering the product vectors into $C$ clusters and calculating transition probabilities between them. After identifying which cluster $p_i$  belongs to (e.g., $p_i \in c_i$), we rank all clusters by their transition probabilities with respect to $c_i$. Then, the products from top clusters are sorted by cosine similarity to $p_i$, where top $K_c$ from each cluster are used as recommendations ($\sum K_c = K$). Example predictions of $\text{bagged-prod2vec-cluster}$ compared to predictions of $\text{bagged-prod2vec}$ are shown in Table~\ref{tab:doc2vec_pred1}. It can be observed that predictions based on the clustering approach are more diverse.

{\bf 4) $\text{user2vec}$} was trained using data set $D_{p}$ where product vectors and user vectors were learned by maximizing log-likelihood proposed in \eqref{user2vec_obj} (see Figure~\ref{fig:user2vec model}). Recommendations for a given user $u_n$ were given by calculating cosine similarity between the user vector $u_n$ and all product vectors, and retrieving the top $K$ nearest products.

{\bf 5) $\text{co-purchase}$.} For each product pair $(p_i, p_j)$ we calculated frequency $F_{(p_i, p_j)}, i=1, \ldots, P, j=1, \ldots, P$, with which product $p_j$ was purchased immediately after product $p_i$. Then, recommendations for a product $p_i$ were given by sorting the frequencies $F_{(p_i, p_j)}, j=1, \ldots, P$, and retrieving the top $K$ products. 

\begin{figure}[t]
\centering
{\includegraphics[width=0.405\textwidth]{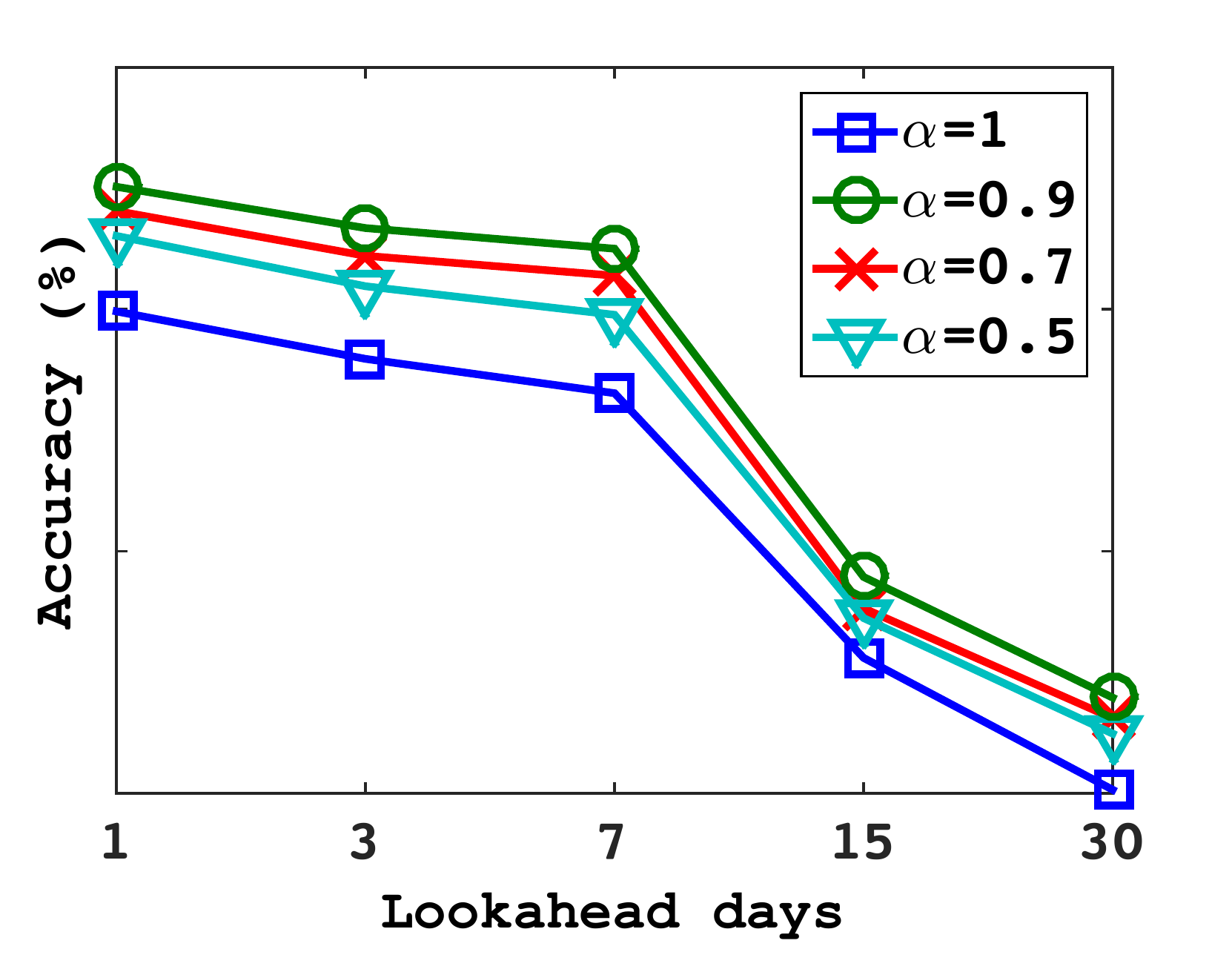}} 
\caption{prod2vec accuracy with different decay values}
\label{fig:decay}
\end{figure}

Since user $u_n$ may have multiple products purchased prior to day $t_d$, separate predictions need to reach a consensus in order to choose the best $K$ products to be shown on that day. To achieve this we propose time-decayed scoring of recommendations, followed by choice of top $K$ products with the highest score. More specifically, given user's products purchased prior to $t_d$ along with their timestamps, $\{(p_{1},t_1),\ldots (p_{U_n},t_{U_n})\}$, for each product we retrieve top $K$ recommendations along with their similarity scores, resulting in the set $\{(p_j, sim_j), j=1, \ldots, K U_n$\}, where $sim$ denotes cosine similarity. Next, we calculate a decayed score for every recommended product,
\begin{equation}\label{score}
d_j =  sim_j \cdot \alpha^{(t_d - t_i)},
\end{equation}
where $(t_d - t_i)$ is a difference in days between current day $t_d$ and the purchase time of product that led to recommendation of  $p_j$, and $\alpha$ is a decay factor. Finally, the decayed scores are sorted in descending order and the top $K$ products are chosen as recommendations for day $t_d$.

{\bf Training details.} Neural language models were trained using a machine with 96GB of RAM memory and 24 cores. Dimensionality of the embedding space was set to $d=300$, context neighborhood size for all models was set to $5$. Finally, we used $10$ negative samples in each vector update. Similarly to the approach in \cite{mikolov2013distributed}, most frequent products and users were subsampled during training. To illustrate the performance of the language models, in Table \ref{tab:prod2vec_pred} we give examples of product-to-product recommendations computed using bagged-prod2vec, where we see that the neighboring products are highly relevant to the query product (e.g., for ``despicable me" the model retrieved similar cartoons).

{\bf Evaluation details.} Similarly to how popular product accuracy was measured, we assumed a daily budget of $K=20$ distinct product recommendations per user. Predictions for day $t_d$ are based on prior purchases from previous days, and we did not consider updating predictions for day $t_d$ using purchases that happened during that day.

{\bf Results.} In the first set of experiments we evaluated performance of prod2vec for different values of decay factors. In Figure~\ref{fig:decay} we show prediction accuracy on test data $D_{p}^{ts}$ when looking $1, 3, 7, 15$, and $30$ days ahead. Initial prod2vec predictions were based on the last user purchase in the training data set $D_{p}$. The results show that discounting of old predictions leads to improved recommendation accuracy, with decay factor of $\alpha=0.9$ being an optimal choice.

\begin{figure}[t]
\centering
{\includegraphics[width=0.4\textwidth]{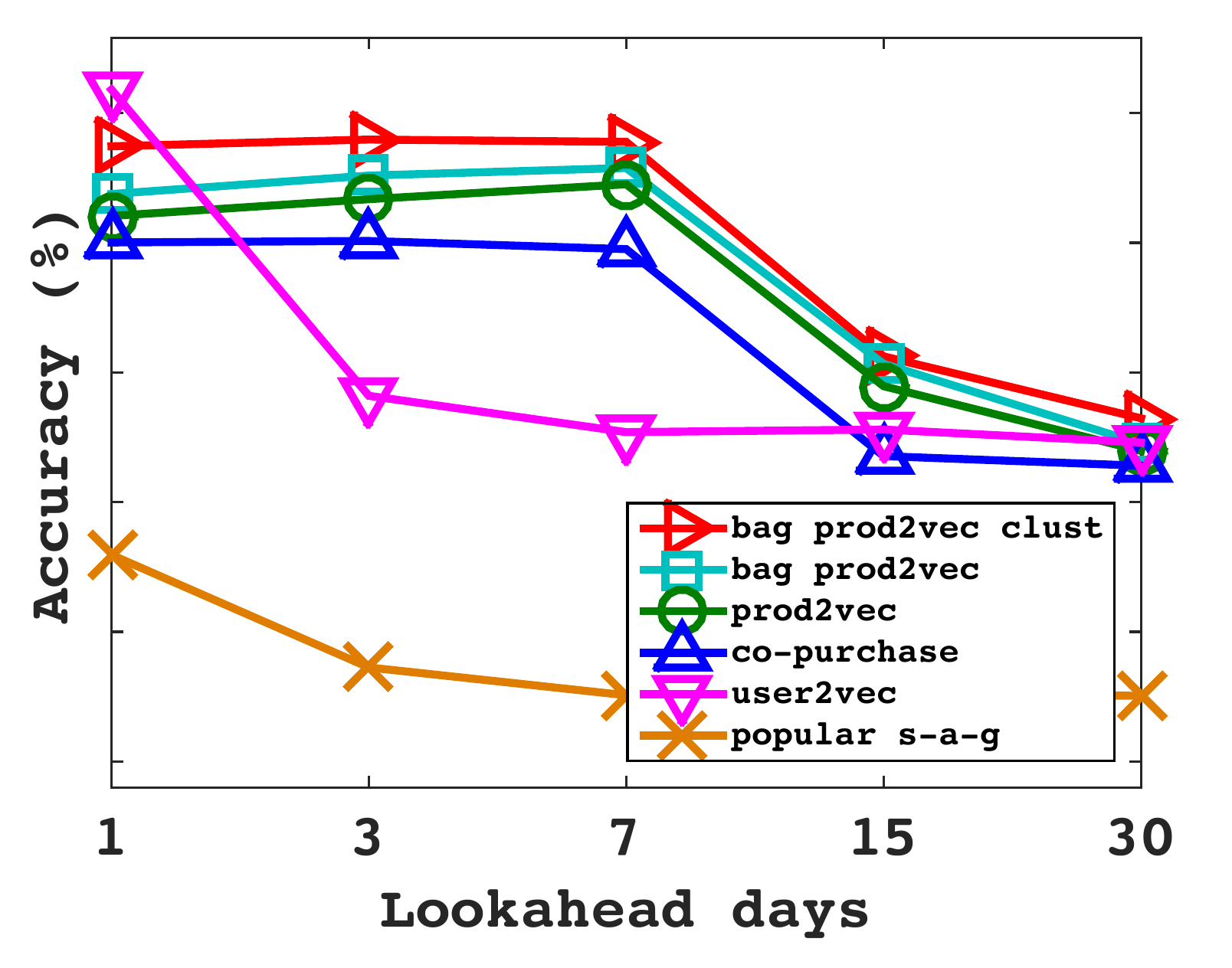}} 
\caption{Prediction accuracy of different algorithms}
\label{fig:pred_results}
\end{figure}

Next, we evaluate different product-to-product and user-to-product methods and compared them to the popular products approach found in Figure \ref{fig:global_cohort} to perform the best, namely recommending popular products computed separately for each (state, age, gender) cohort. Results are summarized in Figure~\ref{fig:pred_results}. We can observe that all neural-based prediction algorithms outperformed method that recommends popular products. Further, even though user2vec model had the best overall accuracy on day 1, its prediction power quickly drops after 3 days. 

On the other hand, variants of the prod2vec model did not exhibit such behavior, and their performance remained steady across the first 7 days. Of all prod2vec models, the bagged-prod2vec-cluster model achieved the best performance, indicating that diversity in predictions leads to better results. Finally, predictions based on co-purchases did not perform as well as neural language models, supporting the ``don't count, predict" claim from \cite{baroni2014don} where the authors suggest that simple co-occurence approaches are suboptimal. 

\subsection{Bucket results}
Following offline evaluation of different product prediction methods, we conducted additional A/B testing experiments on live Yahoo Mail traffic.
We ran two buckets with two different recommendation techniques, both on $5\%$ of Yahoo Mail users. In the first bucket, for users with prior purchases the recommendations were based on the bagged-prod2vec-cluster model, while for users without prior purchases they were based on globally popular products. In the second bucket all users were recommended globally popular products. For purposes of a fair bucket test, both models were retrained and refreshed with the same frequency of $7$ days.

\begin{table}
{\footnotesize
\caption{Results from live A/B testing of product recommendations on Yahoo Mail}
\label{tbl:results_bucket}
\begin{center}
\begin{tabular}{lccc}
 & {\bf Control } & {\bf Popular }  & {\bf Predicted }\\
\rule{0pt}{2.5ex}{\bf Metric} &  {\bf (5\% traffic) } & {\bf (5\% traffic) } & {\bf (5\% traffic) }\\
\hline \hline
\rowcolor{lightgray}
CTR & - & +8.33\% & +9.81 \% \\ 
YR &  & - & +7.63 \% \\
\bottomrule
\end{tabular}
\end{center}
}
\end{table}

\begin{figure}[t]
\centering
{\includegraphics[width=0.45\textwidth]{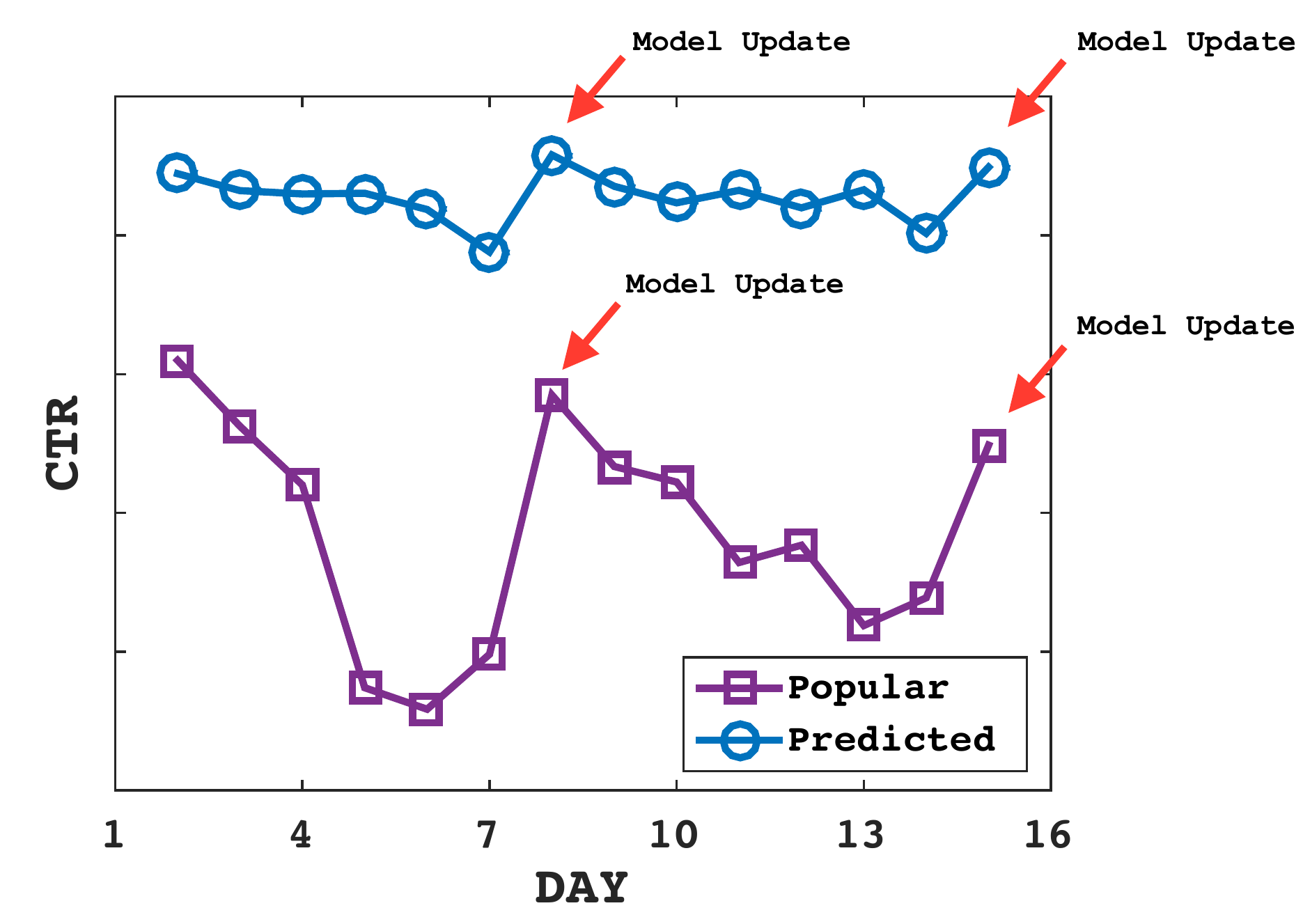}} 
\caption{CTR of predicted versus popular recommendations in live bucket test over time}
\label{fig:p2p_stats_bucket}
\end{figure}

Both test buckets were evaluated against a control bucket in which products ads were replaced with standard ads from Yahoo Ad serving platform. All three buckets had significant amount of users with prior purchases. 

Evaluation was done based on the click-through rate (CTR) computed as $\text{CTR}=\frac{\#\text{clicks}}{\#\text{impressions}}$. In particular, we measured the number of clicks on product ads that occurred after the targeted recommendation, divided by the total number of shown (or impressed) product ads. In the control bucket, we computed the same metric for standard ads. For the two product ad buckets we additionally measured the yield rate, calculated as $\text{YR}=\frac{\#\text{conversions}}{\#\text{impressions}}$, where conversion refers to an actual purchase of the recommended product. Conversions were attributed to impression for up to $48$ hours since the time users clicked or observed a product ad.

The results are presented in Table ~\ref{tbl:results_bucket}. We can make several observations. First, both popular and predicted bucket showed better CTR numbers than the control bucket, indicating that the users prefer product ads over standard ads. Second, the prediction bucket achieved slightly better CTR rates than the popular bucket. Finally, the prediction bucket achieved significantly better YR than the popular bucket. This indicates that many times users click on popular products out of curiosity and do not end up buying the product, whereas clicks on targeted products lead to more purchases as they better capture user interests. Overall, the presented results strongly suggest benefits of the proposed approach for the task of product recommendations.

In addition, in Figure ~\ref{fig:p2p_stats_bucket} we show how CTR rates change over time. Similarly to our offline experiments, it can be observed that popular recommendations become stale much faster than predicted recommendations, as indicated by the steeper CTR drop. They are also more susceptible to novelty bias following model updates, as seen by larger increase in CTR. 
Another useful finding was that $7$ day updates are not sufficient, confirming findings from Figure \ref{fig:global}. 

\section{System Deployment}

In this section we cover the details on our system implementation that led to final product deployment. 

\subsection{Implementation details}

Due to product requirements for near-real-time predictions, we chose to use product-to-product recommendations, with the bagged-prod2vec-cluster model. The model is updated every $5$ days with the most recent purchase data. The product vectors are stored on Hadoop Distributed File System (HDFS), and updated via training procedure  implemented on Hadoop\footnote{https://hadoop.apache.org, accessed June 2015}, where we heavily leverage parallel processing with MapReduce. We form purchase sequences using the newest purchased products extracted from e-mail receipts, and incrementally update the product vectors instead of training them from scratch.

Popular products were used as back-fill for users who do not have prior purchase history. Following the experimental results, we recalculated popular products every $3$ days, with a lookback of $5$ days. More specifically, we extracted $100$ most frequently purchased products for every state-age-gender cohort, and showed them in randomized order to users without purchase history.

We implemented a multi-tier architecture in order to productionize product recommendation capability on Yahoo Mail. For data storage we used a custom, high-performance distributed key-value store (similar to Cassandra\footnote{http://cassandra.apache.org/, accessed June 2015}), which persists user profiles and product-to-product prediction model. The user profile store utilizes user identifier as a key and stores multiple columns representing user prior purchases as values. User profiles are updated hourly  with new products, extracted from e-mail receipts. Each purchase record persists in memory with time-to-live (TTL) set to 60 days, after which period it is discarded.

Product-to-product prediction model is stored in a similar key-value store, where purchased product is used as a key and values are multiple columns each representing predicted products ordered by score. Both user and product-to-product stores are updated in batch without impacting real traffic. Separate process performs all processing of querying user purchases and asynchronously retrieves relevant predictions and offers. It then returns selected product ad back to presentation process implemented in JavaScript and HTML which renders it in the browser. The retrieval process is configured such that it can fetch offers from multiple e-commerce websites (affiliate partners). The system runs with a service-level agreement of 500ms and can be improved further by caching interactions with affiliate partners. 

Given predictions for a certain user, once the user logs into the e-mail client we show a new recommendation after every user action, including clicking on folders, composing e-mail, and searching the inbox. Recommendation are circled in the order of decayed prediction score, as described in Section~\ref{sec:predicted}. The product ads are implemented in the so-called ``pencil" ad position, just above the first e-mail in the inbox (Figure~\ref{fig:prod_ads}).

\section{Conclusions and FUTURE WORK}
We presented the details of a large-scale product recommendation framework that was launched on Yahoo Mail in the form of product ads. We discussed the recommendation methodology, as well as the high-level implementation details behind our system. To perform recommendations, we described how we employed neural language models capable of learning product embeddings for product-to-product predictions, as well as user embeddings for user-to-products predictions. Several variants of the prediction models were tested offline and the best candidate was chosen for an online bucket test. Following the encouraging bucket test results, we launched the system in production. In our ongoing work, we plan to utilize implicit feedback available through ad views, ad clicks, and conversions to further improve the performance of our product recommendation system.

\balance

\bibliographystyle{abbrv}
\bibliography{search_refs}

\end{document}